\pdfoutput=1
\documentclass[sigconf]{acmart}
\usepackage{amsmath}
\usepackage{booktabs}
\usepackage{multirow}
\usepackage{graphicx}
\usepackage{xcolor}
\usepackage{algorithm}
\usepackage{algorithmic}
\usepackage{subcaption}
\usepackage{placeins}
\usepackage{xspace}

\newcommand{\pharmlevel}{phar\-ma\-co\-phore-level\xspace}

\newcommand{\methodname}{{VINCENT}\xspace}
\newcommand{\ReLU}{\mathrm{ReLU}}

\setcopyright{none}
\begin{document}

%% ----- Title -----
\title{VINCENT: Validated Interaction Network \\
for Cross-drug Explanation of Therapeutics}

%% ----- Authors-----
\author{Fan-Sheng Chuang}
\affiliation{
  \institution{North Carolina State University}
  \department{Computer Science}
  \city{Raleigh}
  \country{USA}
}
\email{fchuang2@ncsu.edu}

\author{Xuchen Li}
\affiliation{
  \institution{North Carolina State University}
  \department{Electrical and Computer Engineering}
  \city{Raleigh}
  \country{USA}
}
\email{xli263@ncsu.edu}

\author{Yujing Bian}
\affiliation{
  \institution{North Carolina State University}
  \department{Electrical and Computer Engineering}
  \city{Raleigh}
  \country{USA}
}
\email{ybian3@ncsu.edu}

\author{Kaixiong Zhou}
\affiliation{
  \institution{North Carolina State University}
  \department{Electrical and Computer Engineering}
  \city{Raleigh}
  \country{USA}
}
\email{kzhou22@ncsu.edu}

%% ----- Abstract -----
\begin{abstract}
Drug synergy prediction estimates whether two drugs produce a stronger joint effect than expected from their individual activities. For such predictions to support drug combination discovery, a single synergy score is often not enough: researchers also need to know which molecular regions of the two drugs jointly drive the predicted effect. We study \emph{motif-pair synergy explanation}, whose goal is to identify which pairs of chemically coherent regions, one from each drug, jointly contribute to the predicted synergy. Recent interpretable synergy models expose atom- or substructure-level signals, but their explanation mechanisms are typically built into the predictor architecture, and none validates its cross-drug region scores under repeated perturbation or feeds that evidence back to refine the explanation. A reliable motif-pair explanation should instead be chemically coherent, so its regions are connected motif pairs across the two drugs; perturbation-stable, so scores reproduce across repeated perturbations rather than a single noisy forward pass; and aligned with predictor behavior.

We introduce \methodname{} (\emph{Validated Interaction Network for Cross-drug Explanation of Therapeutics}), a post-training framework that explains a fixed interaction-aware synergy predictor. \methodname{} extracts atom-pair evidence from the predictor's attention and gradient signals, groups atoms into chemically coherent motifs, and validates candidate motif pairs through repeated local perturbation. The validated evidence is fed back to refine the motif assignments, yielding explanations that are chemically coherent, perturbation-stable, and aligned with predictor behavior. On the 25-pair literature-annotated subset, \methodname{} achieves mean motif recall of 0.826 (95\% CI: 0.78--0.87), compared with 0.49--0.66 for baselines. Across all 71 test pairs, its validated interaction scores yield a TP/TN separation of 3.36. These results show that closed-loop perturbation validation recovers literature-supported molecular regions more accurately than existing alternatives while producing cross-drug interaction scores that better reflect predictor behavior.
\end{abstract}

%% ----- CCS / Keywords -----
\begin{CCSXML}
% TODO: fill CCS codes
\end{CCSXML}
\keywords{Drug synergy prediction, motif-pair synergy explanation, graph neural networks, motif substructure learning.}

\maketitle

%% ============================================================
%% §1 INTRODUCTION
%% ============================================================
%% ============================================================
%% §1 INTRODUCTION
%% ============================================================
\section{Introduction}
\label{sec:intro}

% ---- P1: Gap ----
Drug-pair synergy prediction asks whether two drugs act together more effectively than expected from their individual effects. It is central to drug combination discovery across drug repurposing, lead optimization, and antiviral combination screening, including COVID-19 settings~\cite{bobrowski2021sarscov2synergy,wagoner2022hostvirus}. Deep learning models predict drug-pair synergy scores~\cite{preuer2018deepsynergy,jin2021deeplearningcovid,wang2022deepdds,liu2021transynergy}, but most work targets the accuracy of a single scalar score, which does not reveal which molecular regions drive the predicted synergy. We study \emph{motif-pair synergy explanation}: identifying which pairs of chemically coherent regions (motifs), one from each drug, jointly contribute to the predicted synergistic effect. This matters in practice because a medicinal chemist needs trustworthy interaction evidence, not only a numerical prediction.

% ---- P2: Three requirements ----
We argue that a useful synergy explanation should be a validated cross-drug motif-pair interaction matrix satisfying three requirements. First, \textbf{chemical structure coherence}~(R1): each motif is a connected region of the molecular graph, and the explanation encodes cross-drug motif pairs rather than isolated atom scores. Second, \textbf{motif-pair perturbation stability}~(R2): the explanatory effect assigned to a motif pair should be supported by consistent predictor responses across repeated local perturbations of that pair, rather than by a single forward-pass attribution that may be noisy or unrepeatable. Third, \textbf{predictor-explainer alignment}~(R3): across drug pairs, the aggregate validated motif-pair evidence should be positively associated with the predictor's synergy output, so the explanation reflects the predictor's decision behavior.

% ---- P3: Existing methods fall short ----
Existing work does not fully address this task. Existing drug-synergy models~\cite{preuer2018deepsynergy,jin2021deeplearningcovid,wang2022deepdds,liu2021transynergy} primarily optimize combination prediction; although some provide pathway- or substructure-level interpretability, they do not produce perturbation-validated cross-drug molecular-region-pair explanations. General GNN explainers~\cite{ying2019gnnexplainer,yuan2021subgraphx,luo2020pgexplainer,lucic2022cfgnnexplainer} identify important nodes, edges, or subgraphs for an individual input graph, but are not designed for pair-conditioned cross-drug interactions. Recent interpretable synergy models provide richer structural or biological explanations, including molecular substructures~\cite{liu2024sddsynergy,guo2024synergyx,xia2026grafsyn}, cross-drug structural attention~\cite{xin2026deepdrugs}, biological or causal subnetworks~\cite{li2025casynergy,dong2021idsp,dong2023sanepool}, and multi-scale structural representations~\cite{huang2026deepstfsynergy}. However, these explanation mechanisms are typically integrated into their respective predictor architectures rather than designed for post-training analysis of a separately trained model. Moreover, to our knowledge, no existing method closes the loop between explanation and validation by repeatedly estimating the stability of cross-drug region-pair effects under local perturbation and feeding those scores back to update the region partition itself. Table~\ref{tab:related-substructure} summarizes these distinctions.

% ---- P4: VINCENT ----
We propose \methodname{} (\emph{Validated Interaction Network for Cross-drug Explanation of Therapeutics}), a post-training framework for extracting validated cross-drug motif-pair explanations from a fixed interaction-aware drug synergy predictor. The predictor supplies atom-level representations, cross-drug signals, and the predicted synergy score; predictor design itself is not a contribution of this work. \methodname{} uses these internal signals to construct chemically coherent motifs, screens candidate motif pairs, and validates the retained pairs through repeated local perturbation. The validated evidence is fed back to refine the motif assignments, closing the loop between explanation and validation. To our knowledge, this is the first drug-synergy explanation framework in which perturbation-derived validation is part of the explanation-generation loop itself, actively reshaping the learned molecular regions rather than serving only as a post-hoc check. We make four contributions.

% ---- P5: Contributions ----
\begin{enumerate}
  \item \textbf{Problem formulation.} We formulate post-training drug synergy explanation as the recovery of a validated cross-drug motif-pair interaction matrix, operationalized through three testable requirements: chemical coherence, perturbation stability, and predictor alignment (R1--R3), formalized in Section~\ref{sec:problem}.
  \item \textbf{Closed-loop explanation method.} We propose \methodname{}, which combines predictor-derived cross-drug evidence, motif construction, repeated local perturbation validation, and feedback refinement in a closed explanation--validation loop. Validated motif-pair evidence is used not only to score the explanation but also to refine the explanatory regions themselves.
  \item \textbf{Literature-grounded evaluation protocol.} We construct a literature-grounded reference set of 111 \pharmlevel{} molecular-region annotations across 25 literature-supported drug pairs, and use it to evaluate whether explanation methods recover pharmacologically relevant regions. All annotations are traceable to their supporting sources, as documented in Appendix~\ref{app:eval-protocol}.
  \item \textbf{Results and significance.} On the literature-annotated subset of 25 pairs, \methodname{} substantially improves the recovery of literature-supported molecular regions, achieving mean motif recall of 0.826 compared with 0.49--0.66 for the evaluated explanation baselines. Across the full 71-pair test set, its validated interaction scores also show stronger predictor alignment, including a TP/TN separation of 3.36.
\end{enumerate}

%% ============================================================
%% §2 RELATED WORK
%% ============================================================
\section{Related Work}
\label{sec:related}

\paragraph{Drug synergy prediction.}
Drug-synergy models span descriptor- and fingerprint-based neural networks~\cite{preuer2018deepsynergy}, graph-based architectures~\cite{wang2022deepdds,graphsynergy2021}, and more recent dual-view, mechanism-informed, similarity-network, knowledge-graph, and multimodal formulations~\cite{liu2021transynergy,jointsyn2024,lai2026dsimsynergy,li2023gaecds,yan2026kglgansynergy,saeed2026mmdgnn}. Large-scale combination databases such as DrugComb~\cite{zagidullin2019drugcomb} and DrugCombDB~\cite{liu2020drugcombdb} have further accelerated model development; see Abbasi and Rousu~\cite{abbasi2024minireview} for a recent survey. Our experimental predictor builds on the multi-task training and Bliss-based formulation of ComboNet~\cite{jin2021deeplearningcovid} and serves solely as the fixed model to be explained. Predictor architecture and training details are described in Section~\ref{sec:backbone} and Appendix~\ref{app:predictor}.

\paragraph{GNN explainability.}
Attribution methods such as Integrated Gradients~\cite{sundararajan2017axiomatic} estimate input-component contributions by path integration, while attention weights alone are not necessarily faithful explanations~\cite{jain2019attention}. Mask-learning approaches such as GNN\-Explainer~\cite{ying2019gnnexplainer} and PG\-Explainer~\cite{luo2020pgexplainer}, and subgraph search methods such as SubgraphX~\cite{yuan2021subgraphx}, provide complementary strategies for explaining graph predictions. Counterfactual approaches such as CF-GNN\-Explainer~\cite{lucic2022cfgnnexplainer} and cooperative explanation methods~\cite{fang2023cooperative} address the same task from different angles; broader surveys appear in~\cite{jimenezluna2020xai_drug_discovery,vo2022xai_ddi_review,ding2025xai_drug_research}. These methods typically identify important nodes, edges, or subgraphs for an individual input graph, whereas our task requires pair-conditioned explanations over molecular regions from both drugs. CF-GNN\-Explainer, for example, searches for minimal graph perturbations that change a prediction; \methodname{} instead uses repeated local feature perturbations to estimate the magnitude and stability of a cross-drug region-pair effect, without requiring the prediction to flip.

\paragraph{Substructure-level synergy explanation.}
Recent interpretable synergy models provide structural and biological interpretations at several levels. SDDSynergy~\cite{liu2024sddsynergy} learns adaptive molecular substructures, while SynergyX~\cite{guo2024synergyx} uses predefined substructures within a mutual-attention architecture. GraFSyn~\cite{xia2026grafsyn} uses connected graph\-lets, while DeepSTF\-Synergy~\cite{huang2026deepstfsynergy} models atomic, substructural, and global structure. DeepDrugs~\cite{xin2026deepdrugs} maps cross-drug attention signals to pharmacophoric regions, whereas SDCInterpreter~\cite{wang2026sdcinterpreter}, IDSP~\cite{dong2021idsp}, SANEPool~\cite{dong2023sanepool}, and CASynergy~\cite{li2025casynergy} provide mechanism-path or biological-network interpretations. Outside synergy-specific models, Wu et al.~\cite{wu2023sme} use molecular fragmentation and masking for post-hoc single-molecule explanation. These works establish the value of substructure- and mechanism-level interpretation, but synergy-specific approaches are typically integrated into their predictor architectures, while generic post-hoc methods do not repeatedly validate cross-drug region pairs or feed the resulting evidence back into region construction. \methodname{} instead combines post-training motif construction with repeated cross-drug perturbation validation and feedback refinement in a closed explanation--validation loop. Table~\ref{tab:related-substructure} summarizes these distinctions.

\begin{table}[t]
\centering
\caption{Comparison of interpretable and post-hoc approaches relevant to cross-drug motif-pair explanation. \emph{Post-hoc}: applicable to an independently trained, fixed predictor. \emph{Rep.\ valid.}: the same cross-drug region pair is repeatedly perturbed to estimate a stable interaction effect. \emph{Feedback}: validated evidence is used to update the explanatory regions. \checkmark/-- denote yes/no.}
\vspace{-10pt}
\label{tab:related-substructure}
\small
\setlength{\tabcolsep}{4pt}
\begin{tabular}{@{}p{2.15cm}p{2.0cm}ccc@{}}
\toprule
Method & Expl.\ unit & Post-hoc & Rep.\ valid. & Feedback \\
\midrule
SDDSynergy~\cite{liu2024sddsynergy}             & Learned substruct.   & -- & -- & -- \\
SynergyX~\cite{guo2024synergyx}                 & Predef.\ substruct.  & -- & -- & -- \\
GraFSyn~\cite{xia2026grafsyn}                   & Graphlet substruct.  & -- & -- & -- \\
DeepSTF\-Synergy~\cite{huang2026deepstfsynergy}  & Multi-scale struct.  & -- & -- & -- \\
DeepDrugs~\cite{xin2026deepdrugs}               & Cross-drug attn.     & -- & -- & -- \\
SDC\-Interpreter~\cite{wang2026sdcinterpreter}   & Mechanism paths      & -- & -- & -- \\
CASynergy~\cite{li2025casynergy}                 & Causal / gene net.   & -- & -- & -- \\
IDSP~\cite{dong2021idsp}                         & Signaling net.       & -- & -- & -- \\
SANEPool~\cite{dong2023sanepool}                 & Gene subnetwork      & -- & -- & -- \\
Wu et al.~\cite{wu2023sme}                      & Predef.\ fragments   & \checkmark & -- & -- \\
PG\-Explainer~\cite{luo2020pgexplainer}          & Learned edge mask    & \checkmark & -- & -- \\
CF-GNN\-Explainer~\cite{lucic2022cfgnnexplainer} & CF edges             & \checkmark & -- & -- \\
\methodname{} (ours)                             & Multi-view motifs    & \checkmark & \checkmark & \checkmark \\
\bottomrule
\end{tabular}
\end{table}

%% ============================================================
%% §3 PRELIMINARIES (~0.5 page)
%% ============================================================
\section{Preliminaries}
\label{sec:prelim}

\subsection{Problem Setting}
\label{sec:problem}
\noindent\textbf{Drug synergy prediction.}
Consider a drug pair $(A,B)$ with $N_A$ and $N_B$ atoms, respectively. Each drug is represented as a molecular graph whose nodes are atoms and whose edges are chemical bonds. A structure learning model obtains atom-level representations from this molecular input, for example using a graph neural network (GNN) operating directly on the drug graph. We denote the resulting atom-level representations by $H_A \in \mathbb{R}^{N_A \times d_h}$ and $H_B \in \mathbb{R}^{N_B \times d_h}$. The single-drug activities $P_A$ and $P_B$ are inferred by feeding each drug representation into a prediction head, while the combination activity $P_{AB}$ is inferred by feeding the pair-conditioned representations into a combination head. Following ComboNet~\cite{jin2021deeplearningcovid}, the Bliss independence baseline is $P_{\mathrm{bliss}} = P_A + P_B - P_A P_B$, and the predicted synergy score is
\begin{equation}
  s_{AB} = P_{AB} - P_{\mathrm{bliss}}.
  \label{eq:synergy-score}
\end{equation}
$P_{\mathrm{bliss}}$ is the combined activity expected if the two drugs acted independently under Bliss independence. Subtracting this baseline from the predicted combination activity $P_{AB}$ measures the gain beyond independent action. Thus, $s_{AB}>0$ indicates positive excess above the Bliss baseline; in our predictor, a pair is classified as synergistic when this excess exceeds 0.5, i.e., $s_{AB}>0.5$.

\noindent\textbf{Motif-pair synergy explanation.}
\emph{Given} a fixed synergy predictor $f$ and a drug pair $(A,B)$, the synergy explanation is to \emph{produce} a validated interaction strength matrix
$R = [r_{kl}] \in \mathbb{R}^{K_A \times K_B}$
over learned motif pairs, where $K_A$ and $K_B$ denote the numbers of motifs identified in the two drugs. Here a motif is a chemically coherent molecular region, and $r_{kl}$ is the validated interaction score between the $k$-th motif of drug~$A$ and the $l$-th motif of drug~$B$, quantifying their joint contribution to the predictor's synergy output. Unlike the scalar $s_{AB}$, $R$ localizes the cross-drug evidence to specific molecular regions. The explanation must satisfy:
\begin{enumerate}
  \item[\emph{(R1)}] \textbf{Chemical structure coherence:} each motif is a connected region of the molecular graph, and $R$ encodes cross-drug motif pairs rather than isolated atom scores.
  \item[\emph{(R2)}] \textbf{Motif-pair  perturbation stability:} each score $r_{kl}$ should be supported by consistent predictor responses across repeated perturbations of the motif pair, rather than by a potentially noisy or unrepeatable single forward pass.
  \item[\emph{(R3)}] \textbf{Predictor-explainer alignment:} across drug pairs, the aggregate interaction strength $\sum_{k,l} r_{kl}$ should be positively associated with the predicted synergy score $s_{AB}$, so the explanation tracks the predictor's own decision behavior.
\end{enumerate}

\subsection{Fixed Predictor Interface}
\label{sec:backbone}
\methodname{} operates on a separately trained interaction-aware predictor that exposes atom-level representations and cross-drug association signals. In our implementation, molecular representations combine a 2D message-passing encoder with geometry-aware 3D representations. The fused representations are then conditioned through a bidirectional atom-level cross-attention module, which exposes an atom-pair association matrix $\hat{A} \in \mathbb{R}^{N_A \times N_B}$, where $\hat{A}_{ij}$ captures the predictor's learned association between atom $i$ of drug~$A$ and atom $j$ of drug~$B$. Separate prediction heads estimate $P_A$, $P_B$, and $P_{AB}$; the synergy score $s_{AB}$ is computed via Eq.~\eqref{eq:synergy-score}. Exact encoder, 2D/3D fusion, and training details are provided in Appendix~\ref{app:predictor}.

The predictor is trained using the drug-target interaction, single-agent activity, and combination objectives adopted from ComboNet. After training, all predictor parameters are frozen. Before explanation, we perform a one-time mask-aware calibration in which only the neutral mask embedding and local reconditioning operator are learned, so that the local feature masking used during perturbation-based validation does not introduce an unfamiliar input pattern. The calibrated perturbation components are then frozen as well; no predictor parameter is updated during calibration or \methodname{} explanation. Calibration details are provided in Appendix~\ref{app:predictor}.

%% ============================================================
%% §4 METHOD (~3 pages)
%% ============================================================
\section{Method}
\label{sec:method}

\methodname{} uses the fixed predictor only as a source of atom-level evidence and prediction outputs, and explains its synergy score by identifying cross-drug molecular-region pairs whose effects remain stable under repeated perturbation. The framework proceeds in four phases. Phases~1--2 extract and organize predictor evidence for motif discovery, while Phases~3--4 form an iterative assignment--validation loop: motifs are constructed, candidate motif pairs are perturbation-validated, and the validated evidence is fed back to refine the next assignment. The evidence map and two base affinity views remain fixed; only motif assignments, validated scores, and feedback affinity evolve across iterations.

\begin{figure*}[t]
  \centering
  \includegraphics[
  width=1\textwidth,
  trim=0 0.2cm 0 0,
  clip]
  {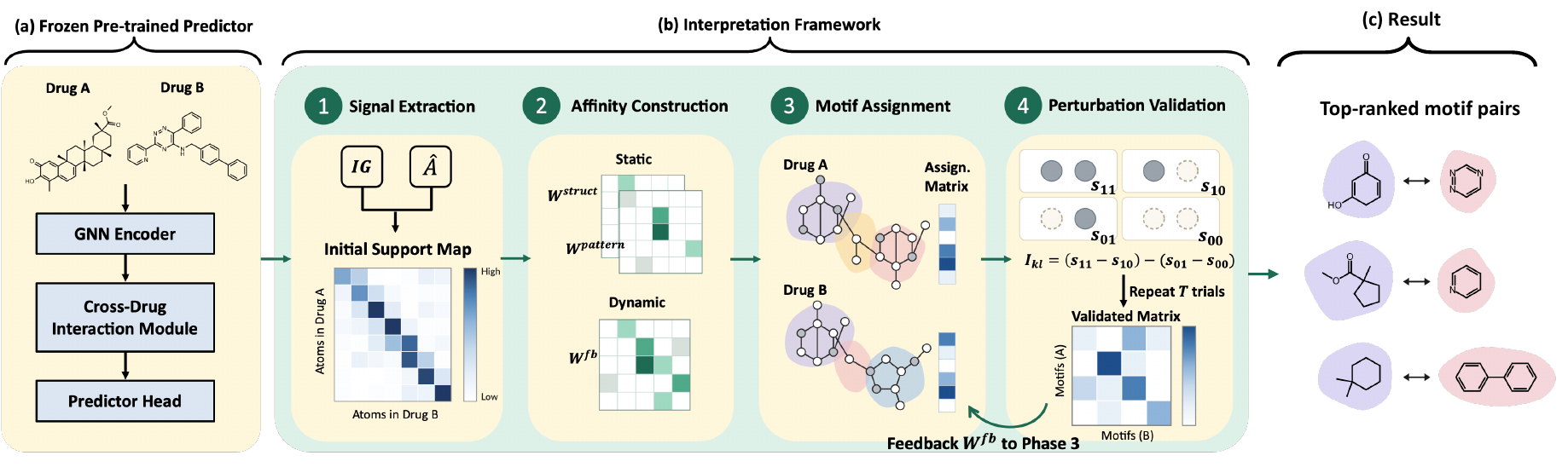}
  \vspace{-20pt}
  \caption{Overview of \methodname{}: the fixed predictor (left) provides atom-level representations and cross-drug signals. Phases~1--2 prepare the loop's inputs; Phases~3--4 form the assignment--validation loop with iterative feedback, yielding the top-ranked cross-drug motif pairs (right). Matrices and molecular fragments are illustrative.}
  \label{fig:pipeline}
\end{figure*}

%% ----- §4.1 -----
\subsection{Phase 1: Cross-drug Signal Extraction}
\label{sec:phase1}

The goal of Phase~1 is to extract atom-pair evidence from the fixed predictor of where the two drugs jointly contribute to the predicted synergy. We start from the cross-drug association matrix $\hat{A} \in \mathbb{R}^{N_A \times N_B}$, where $\hat{A}_{ij}$ is the predictor's learned association between atom $i$ of drug~$A$ and atom $j$ of drug~$B$. This matrix identifies where the predictor establishes cross-drug associations at the atom level, but association alone does not guarantee relevance to the synergy score. We therefore also compute pairwise Integrated Gradients~\cite{sundararajan2017axiomatic}, yielding $IG \in \mathbb{R}^{N_A \times N_B}$, where $IG_{ij}$ measures the contribution of atom pair $(i,j)$ to $s_{AB}$. Integrated Gradients are directly tied to the prediction target but can be noisy in isolation.

We define the atom-pair evidence map as the positive intersection of these two signals:
\begin{equation}
  M = \ReLU(\hat{A}) \odot \ReLU(IG),
  \label{eq:m0}
\end{equation}
where $\odot$ denotes element-wise multiplication. The $\ReLU$ operations retain only positive values; the product suppresses atom pairs supported by only one signal. A large $M_{ij}$ indicates that atom pair $(i,j)$ is both associated by the predictor and positively involved in increasing $s_{AB}$. The matrix $M$ is computed once and remains fixed throughout the explanation process.

%% ----- §4.2 -----
\subsection{Phase 2: Within-drug Affinity Construction}
\label{sec:phase2}

The matrix $M \in \mathbb{R}^{N_A \times N_B}$ identifies important atom pairs across the two drugs, but motif discovery requires a within-molecule relation: which atoms should be grouped together. Phase~2 therefore constructs three complementary affinity matrices for each drug. For drug~$A$, each $W_A^V \in \mathbb{R}^{N_A \times N_A}$, where $W_A^V[i,j]$ measures how strongly atoms $i$ and $j$ should share a motif under view $V$; drug~$B$ is treated analogously. The three views encode distinct grouping cues: chemical locality, partner-conditioned predictor behavior, and perturbation-validated feedback.

The \emph{structural view} $W_A^{\mathrm{struct}}$ uses a Gaussian decay of shortest-path distance in the molecular graph, providing a chemical-locality prior for coherent regions. The \emph{interaction-pattern view} $W_A^{\mathrm{pattern}}$ favors nearby, sufficiently active atoms with similar evidence profiles $M[i,:]$ toward drug~$B$. Thus, structural affinity asks whether two atoms are plausible local motif members, whereas pattern affinity asks whether the fixed predictor uses them similarly with respect to the partner drug. These two base views are computed once---the former from molecular structure and the latter from $M$---and remain fixed.

The \emph{feedback view} $W_A^{\mathrm{fb}}$ supplies information unavailable before validation. It is initialized to zero; after Phase~4 produces validated motif-pair scores, that evidence is projected back to atoms so that nearby atoms with similar validated interaction profiles receive higher affinity. This is the only dynamic view, allowing perturbation-stable evidence to reshape the next motif assignment without modifying the original evidence map $M$. Before Phase~3, each affinity is converted into a normalized Laplacian. Exact operators are given in Appendix~\ref{app:method-details}.

No single view provides all three cues: structural affinity is chemically
grounded but partner-agnostic, whereas interaction-pattern affinity is
partner-conditioned but still derived from the unvalidated evidence map
$M$. The feedback view supplies the missing post-validation signal, so the
three views respectively capture structure, initial predictor evidence,
and validated interaction evidence.

%% ----- §4.3 -----

\subsection{Phase 3: Constrained Motif Assignment}
\label{sec:phase3}

Phase~3 groups each drug's atoms into chemically coherent motifs by learning a soft assignment matrix $S_A \in \mathbb{R}^{N_A \times K_A}$, where $S_A[i,k] \geq 0$ denotes the soft membership of atom $i$ in motif $k$ and each row sums to one. The same process applies to drug~$B$. The assignment is obtained by minimizing:
\begin{equation}
  \mathcal{L}(S_A) = \sum_{V \in \{\mathrm{struct},\, \mathrm{pattern},\, \mathrm{fb}\}} \lambda_V \, \mathrm{Tr}\!\bigl(S_A^\top L_A^V S_A\bigr) + \mathcal{R}(S_A),
  \label{eq:assignment-loss}
\end{equation}
where $L_A^V$ is the normalized Laplacian of affinity view $V$. The three smoothness terms encourage atoms to share a motif when they are chemically local, exhibit similar partner-conditioned interaction patterns, or receive similar validated feedback. The regularizer $\mathcal{R}$ combines entropy regularization, which prevents premature collapse of the soft assignment, with a minimum-mass penalty that discourages tiny clusters.

After optimization, each atom is assigned to its highest-weight motif: $\hat{k}(i) = \arg\max_k S_A[i,k]$. We then enforce graph connectivity on the hard assignment so that each resulting motif forms a connected molecular region. A ring-completion step subsequently absorbs remaining ring atoms when a motif already contains at least half of a ring system, preventing chemically indivisible rings from being split. The resulting motifs provide the candidate regions tested in Phase~4.

%% ----- §4.4 -----
\subsection{Phase 4: Perturbation-Based Validation and Feedback}
\label{sec:phase4}

The goal of Phase~4 is to validate whether candidate motif pairs from the two drugs induce stable joint effects in the fixed predictor, and to feed the validated evidence back to improve the next motif assignment. We describe the four steps of this phase in order: screening, perturbation, scoring, and feedback.

\noindent\textbf{Candidate screening.}
We first screen motif pairs by aggregating the fixed evidence $M$ over the current soft assignments $S_A, S_B$ from Phase~3. The coarse score between motif $k$ of drug~$A$ and motif $l$ of drug~$B$ is:
\begin{equation}
  a_{kl} = S_A[:,k]^\top \, M \, S_B[:,l].
  \label{eq:coarse-screening}
\end{equation}
This score aggregates atom-pair evidence weighted by motif membership and is used only to rank candidate pairs; it is not the final interaction score. Only the highest-ranked candidates proceed to perturbation validation.

\noindent\textbf{Repeated local perturbation.}
For each retained motif pair $(k,l)$, we perform $T$ local perturbation trials. Each trial masks a different local subset and fraction of the atoms in the two motifs, producing multiple nearby perturbed realizations of the same motif pair. Masked atom features are replaced by a learned neutral embedding, after which nearby representations are locally reconditioned while the molecular topology remains unchanged. This neutral intervention is supported by the one-time mask-aware calibration described in Section~\ref{sec:backbone}, which reduces distribution shift during perturbation evaluation. Details of the masking and reconditioning operators are in Appendix~\ref{app:perturbation-details}.

\noindent\textbf{Pairwise interaction effect.}
For trial $t$, let $s_{11}^{(t)}$, $s_{10}^{(t)}$, $s_{01}^{(t)}$, and $s_{00}^{(t)}$ denote the predictor's synergy output when both sampled regions are retained, only the region from drug~$A$ is retained, only the region from drug~$B$ is retained, or both are masked, respectively. The pairwise interaction effect is the second-order finite difference:
\begin{equation}
  I_{kl}^{(t)} = \bigl(s_{11}^{(t)} - s_{10}^{(t)}\bigr) - \bigl(s_{01}^{(t)} - s_{00}^{(t)}\bigr).
  \label{eq:interaction-effect}
\end{equation}
A positive $I_{kl}^{(t)}$ indicates a joint effect that cannot be explained by the two local perturbations independently.

\noindent\textbf{Stable interaction score.}
Across the $T$ trials, we summarize the interaction-effect distribution by its mean $\mu_{kl}$, variability $\sigma_{kl}$, activation frequency $q_{kl}$, and positive-effect frequency $p_{kl}^{+}$. The validated interaction score is:
\begin{equation}
  r_{kl} = \mathrm{softplus}\!\left(\frac{\mu_{kl}}{\sigma_{kl}+\epsilon}\right) \cdot \max(0,\, 2p_{kl}^{+}-1) \cdot q_{kl},
  \label{eq:stable-interaction-score}
\end{equation}
which is high only when the effect is large relative to its variability, directionally consistent, and repeatedly activated. The matrix $R = [r_{kl}]$ is the validated interaction matrix at this outer-loop state.

\noindent\textbf{Feedback refinement.}
The validated matrix $R$ completes one half of the closed loop; the other half projects this evidence back into the motif assignment for the next iteration. The feedback operator $\Phi_{\mathrm{fb}}$ connects nearby atoms within the same drug when their validated interaction profiles toward the partner drug are similar. Atoms that repeatedly participate in the same validated cross-drug effects are thereby encouraged to form a common motif in the next iteration. The feedback affinity is smoothed across iterations via exponential moving average to avoid abrupt assignment oscillations:
\begin{equation}
  W_{\mathrm{fb}}^{(t+1)} = \mathrm{EMA}\!\left(W_{\mathrm{fb}}^{(t)},\; \Phi_{\mathrm{fb}}\!\left(S^{(t)}, R^{(t)}\right)\right).
  \label{eq:feedback-update}
\end{equation}
Starting from $W_{\mathrm{fb}}^{(0)} = 0$, each outer-loop iteration alternates between motif assignment (Phase~3), perturbation validation, and feedback update. We use three outer-loop iterations by default; longer runs are used only for convergence analysis (Appendix~\ref{app:sensitivity}). The original evidence map $M$ is never modified by feedback; validation therefore reshapes how the fixed predictor evidence is grouped into motifs, rather than creating new evidence.

\begin{table}[t]
\centering
\caption{Key hyperparameter settings.}
\vspace{-10pt}
\label{tab:hyperparams}
\small
\begin{tabular}{lll}
\toprule
Parameter & Default & Role \\
\midrule
Target motif size $c$ & 6 & Motif granularity \\
Perturbation trials $T$ & 16 & Perturbation stability \\
Outer-loop iterations & 3 & Refinement depth \\
Candidate screening & top 30\%, min 3, max 20 & Pair-selection scope \\
\bottomrule
\end{tabular}
\end{table}

%% ============================================================
%% §5 EXPERIMENTS (~2.5-3 pages)
%% ============================================================
\section{Experiments}
\label{sec:exp}

We evaluate \methodname{} through motif coverage against independently annotated pharmacophore regions, predictor alignment with synergy behavior, and ablation and sensitivity analyses of individual components.

\subsection{Experimental Setup}
\label{sec:exp-dataset}

\noindent\textbf{Dataset.}
We use the SARS-CoV-2 drug-combination benchmark released with ComboNet~\cite{jin2021deeplearningcovid}, which pairs antiviral compounds and measures their combination activity. After canonicalizing molecular representations and deduplicating unordered pairs, the final splits contain 88 training, 19 validation, and 71 test pairs. We adopt the same auxiliary training data (drug--target interaction, single-agent activity, and HIV combination data) as in the original work; all explanation experiments are conducted exclusively on the SARS-CoV-2 combination pairs. \methodname{} is run on all 71 test pairs.

\noindent\textbf{Literature-supported reference set.}
Evaluating motif coverage requires knowing which molecular regions are pharmacologically relevant for each drug pair. Starting from the 71 test pairs, we retain 25 for which published pharmacological sources provide sufficient mechanistic detail to identify specific molecular regions in each drug. For each retained pair, we manually map literature-identified regions to atom indices, producing 111 pharmacophore-level motif annotations in total (averaging 4.4 annotated regions per drug pair, counting both drugs). All annotations are constructed independently of model output; the full construction procedure and source traceability are described in Appendix~\ref{app:eval-protocol}.

\noindent\textbf{Implementation details.}
The predictor parameters are frozen after training and remain fixed during the one-time mask-aware calibration described in Section~\ref{sec:backbone}; only the perturbation-interface components are calibrated. Unless otherwise specified, \methodname{} uses a target motif size of $c{=}6$ atoms, $T{=}16$ perturbation trials per retained motif pair, top-30\% candidate screening (minimum~3, maximum~20 pairs), and 3 outer-loop iterations. The outer-loop depth was selected on validation/pilot data before test-set evaluation.

\noindent\textbf{Baselines.}
We compare \methodname{} against four groups under the same fixed predictor and evaluation protocol. \emph{Attribution baselines} use Integrated Gradients or cross-drug attention to score individual atoms without motif grouping. \emph{External single-graph explainers}---GNN\-Explainer~\cite{ying2019gnnexplainer}, Subgraph\-X~\cite{yuan2021subgraphx}, PG\-Explainer~\cite{luo2020pgexplainer}, and CF-GNN\-Explainer~\cite{lucic2022cfgnnexplainer}---explain each drug separately, with scores converted to candidate regions under the common matching protocol. \emph{Controlled baselines} cluster atoms using either signal and apply the same perturbation-based validation without iterative feedback, isolating the grouping step. The \emph{lower bound} is a random connected substructure of matched size. To keep the comparison controlled, all methods explain the same frozen predictor, and their outputs are converted into molecular regions before evaluation under the same region-matching protocol and size constraints. Thus, differences in coverage reflect the explanation strategy rather than changes in the prediction model or
evaluation rule. Adaptation details appear in Appendix~\ref{app:baseline-setup}.

%% ============================================================
\subsection{Predictor Adequacy and Evaluation Protocol}
\label{sec:exp-protocol}

\noindent\textbf{Predictor adequacy.}
Because post-training explanation is meaningful only for an informative prediction target, we first verify the fixed predictor's predictive adequacy. It achieves a test ROC-AUC of 0.85, compared with 0.82 for the original ComboNet, 0.80 for DeepDDS, 0.68 for DeepSynergy, and 0.62 for a random forest baseline. Full predictor comparison details appear in Appendix~\ref{app:predictor}.

\noindent\textbf{Motif-coverage evaluation.}
We run each explanation method on the 25 annotated test pairs and compare predicted molecular regions against the 111 literature-supported pharmacophore annotations. The evaluation is per reference motif: for each annotated region, we select the best-matching predicted region from a single cluster or a union of at most two bond-adjacent clusters, subject to the size cap $|P| \leq \max(|L|+3,\lceil1.4|L|\rceil)$ to prevent inflated scores. We report recall, precision, Jaccard overlap, and hit rate at recall $\geq 0.7$ (the fraction of the 111 reference motifs for which the best-matching predicted region achieves recall of at least 0.7). All confidence intervals are from pair-level bootstrap (2{,}000 resamples). Robustness analyses appear in Appendix~\ref{app:robustness}.

\noindent\textbf{Predictor-alignment evaluation.}
We additionally assess whether the validated interaction scores reflect the predictor's synergy behavior. This evaluation uses all 71 test pairs, not only the annotated subset. For each pair, Phase~4 produces a validated interaction matrix $R$; we sum the validated interaction scores over all motif pairs into a total explanation strength $\sum_{k,l} r_{kl}$ and compute the Pearson correlation with the predicted synergy score $s_{AB}$. We also report the TP/TN separation ratio: the mean of the highest single validated interaction score $\max_{k,l} r_{kl}$ for predictor true positives (pairs correctly predicted as synergistic by the fixed predictor) divided by that for predictor true negatives (pairs correctly predicted as non-synergistic), testing whether the explanation scores discriminate the two groups.

%% ============================================================
\subsection{Motif Coverage Results}
\label{sec:exp-coverage}

\begin{table*}[t]
\centering
\caption{Main explanation results. Recall, precision, Jaccard, and HR$\geq$0.7 are evaluated on the 25-pair literature-annotated subset (111 reference motifs). Predictor-alignment metrics $\rho(s_{AB},\sum r)$ and TP/TN separation are evaluated on all 71 test pairs. Coverage confidence intervals are from pair-level bootstrap with 2{,}000 resamples.}
\vspace{-10pt}
\label{tab:main-results}
\small
\setlength{\tabcolsep}{4pt}
\begin{tabular}{llcccccc}
\toprule
 & Method & Recall & Precision & Jaccard &
 HR$\geq$0.7 & $\rho(s_{AB},\sum r)$ & TP/TN Sep. \\
\midrule
\multirow{2}{*}{Attribution}
 & Integrated Gradients Saliency
 & 0.498 & 0.344 & 0.288 & 28.8\% & --- & --- \\
 & Cross-Drug Attention Saliency
 & 0.488 & 0.338 & 0.287 & 30.6\% & --- & --- \\
\midrule
\multirow{4}{*}{External}
 & GNNExplainer
 & 0.581 & 0.476 & 0.394 & 39.2\% & --- & --- \\
 & SubgraphX
 & 0.661 & 0.536 & 0.443 & 48.9\% & --- & --- \\
 & PGExplainer
 & 0.552 & 0.451 & 0.371 & 40.8\% & --- & --- \\
 & CF-GNNExplainer
 & 0.486 & 0.397 & 0.311 & 26.7\% & --- & --- \\
\midrule
Lower bound
 & Random Substructure
 & 0.347 & 0.263 & 0.184 & 14.3\% & --- & --- \\
\midrule
\multirow{2}{*}{Controlled}
 & IG Clust.+Pert.
 & 0.586 & 0.539 & 0.392 & 18.0\% & 0.043 & 1.49 \\
 & ATT Clust.+Pert.
 & 0.646 & 0.599 & 0.458 & 38.7\% & $-$0.238 & 0.48 \\
\midrule
\multirow{2}{*}{Ours}
 & No Feedback
 & 0.724 & 0.681 & 0.568 & 54.9\% & 0.352 & 1.95 \\
 & \methodname{} (Full)
 & \textbf{0.826} & \textbf{0.790} & \textbf{0.689}
 & \textbf{76.6\%} & \textbf{0.423} & \textbf{3.36} \\
\bottomrule
\end{tabular}
\end{table*}

\begin{figure*}[t]
    \centering
    \includegraphics[width=0.78\textwidth]{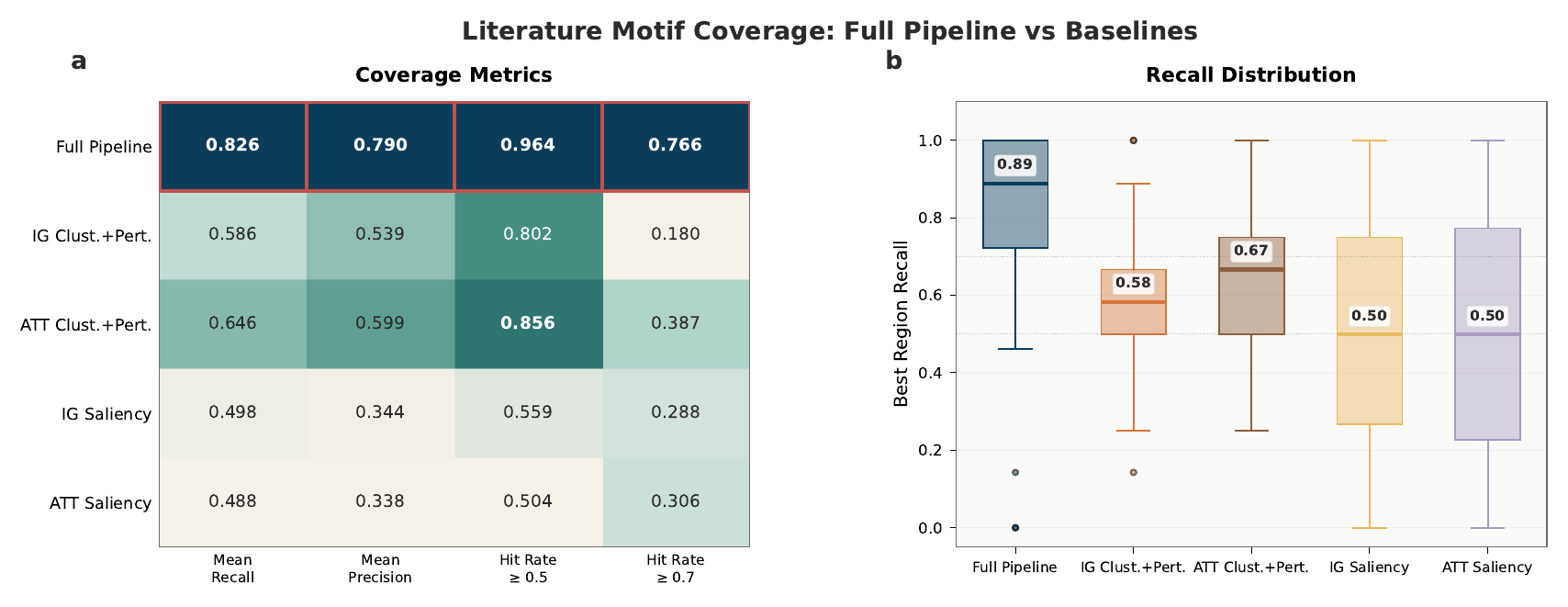}
    \caption{Literature motif coverage: (a)~coverage metrics across methods; (b)~distribution of best-region recall across motifs.}
    \label{fig:motif-coverage}
\end{figure*}

We test whether \methodname{}'s learned motifs overlap with independently annotated pharmacophore regions, indicating pharmacologically meaningful structure. Table~\ref{tab:main-results} and Figure~\ref{fig:motif-coverage} present the comparison.

\textbf{\methodname{} achieves the strongest literature motif coverage among all evaluated methods,} with mean recall of 0.826 (95\% CI: 0.782--0.868), precision of 0.790, Jaccard of 0.689, and a 76.6\% hit rate at recall $\geq 0.7$. Atom-level attribution baselines reach only about 0.49 recall, consistent with their lack of motif construction. External single-graph explainers improve to 0.49--0.66 by producing connected subgraphs, but explain each drug independently of its partner. The controlled single-signal clustering baselines further reach 0.586 and 0.646 under the same perturbation validation. The remaining improvement to 0.826 supports the benefit of combining multi-view assignment with iterative feedback.

%% ============================================================
\subsection{Synergy Discrimination}
\label{sec:exp-discrimination}

Motif coverage asks whether the right molecular regions are recovered; predictor alignment asks the complementary question of whether their validated interaction scores track the fixed predictor's synergy behavior. We evaluate this across all 71 test pairs using the aggregate and top-interaction measures defined in Section~\ref{sec:exp-protocol}.

\textbf{\methodname{}'s validated interaction scores align with the predictor's synergy behavior.} The Pearson correlation between $s_{AB}$ and total validated interaction strength $\sum_{k,l} r_{kl}$ is 0.423. The mean top interaction score is 4.82 for predictor true positives and 1.43 for true negatives, yielding a TP/TN separation of 3.36. In contrast, IG Clust.+Pert.\ yields correlation 0.043 and separation 1.49, while ATT Clust.+Pert.\ yields $-$0.238 and 0.48. Thus, \methodname{}'s validated scores are substantially more associated with the fixed predictor's decision behavior than those of the controlled baselines.

%% ============================================================
\subsection{Ablation and Sensitivity Analysis}
\label{sec:exp-ablation}

The full method combines two ingredients beyond standard single-signal clustering: multi-view affinities for grouping, and iterative feedback from validation back into assignment. We isolate each contribution using Table~\ref{tab:main-results}.

\textbf{Iterative feedback is the largest contributor to the improvement beyond the initial multi-view assignment.} Removing the feedback loop while keeping the multi-view assignment (the ``No Feedback'' variant) reduces recall from 0.826 to 0.724 and TP/TN separation from 3.36 to 1.95, with the synergy correlation dropping from 0.423 to 0.352. This shows that feeding perturbation-validated evidence back into motif assignment materially changes the recovered regions and improves their alignment with predictor behavior.

\textbf{Multi-view assignment improves motif recovery even without feedback.} Even without feedback, the multi-view variant achieves 0.724 recall, compared with 0.586 for IG Clust.+Pert.\ and 0.646 for ATT Clust.+Pert. Because all three variants use the same perturbation-based validation, this comparison directly isolates the grouping step: combining partner-conditioned interaction patterns with molecular locality produces better initial motif candidates than either signal alone.

\textbf{Hyperparameter sensitivity.} Table~\ref{tab:sensitivity_main} reports the effect of varying one parameter at a time. Target motif size is stable between five and eight atoms per motif; substantially coarser clusters reduce coverage. Increasing perturbation trials from 16 to 32 changes recall by only 0.002 and TP/TN separation by 0.08, indicating that the aggregate metrics are stable beyond the default setting. Performance rises rapidly through the first three outer-loop iterations and then stabilizes; complete results for the five hyperparameter sweeps and the early outer-loop convergence analysis appear in Appendix~\ref{app:sensitivity}.

\begin{table*}[t]
\centering
\caption{Sensitivity to target motif size $c$, perturbation trials $T$, and outer-loop refinement. Bold marks the default $c$ and $T$; the default outer-loop depth is 3. $\rho$: Pearson correlation between predicted synergy and total validated interaction strength.}
\vspace{-10pt}
\label{tab:sensitivity_main}
\small
\setlength{\tabcolsep}{5pt}
\begin{tabular}{llcccccc}
\toprule
Parameter & Setting & Recall & Precision & Jaccard &
HR$\geq$0.7 & $\rho$ & TP/TN \\
\midrule
\multirow{5}{*}{Target size $c$}
 & 4  & .791 & .766 & .641 & .69 & .404 & 3.05 \\
 & 5  & .823 & .788 & .686 & .75 & .418 & 3.28 \\
 & \textbf{6}
      & \textbf{.826} & \textbf{.790} & \textbf{.689}
      & \textbf{.77} & \textbf{.423} & \textbf{3.36} \\
 & 8  & .802 & .741 & .632 & .71 & .409 & 3.14 \\
 & 10 & .768 & .699 & .582 & .63 & .392 & 2.88 \\
\midrule
\multirow{4}{*}{Perturbation trials $T$}
 & 4  & .812 & .777 & .668 & .73 & .331 & 2.38 \\
 & 8  & .821 & .785 & .681 & .75 & .401 & 3.02 \\
 & \textbf{16}
      & \textbf{.826} & \textbf{.790} & \textbf{.689}
      & \textbf{.77} & \textbf{.423} & \textbf{3.36} \\
 & 32 & .824 & .791 & .690 & .77 & .431 & 3.44 \\
\midrule
\multirow{5}{*}{Outer-loop state}
 & 0 & .724 & .681 & .568 & .55 & .352 & 1.95 \\
 & 1 & .796 & .758 & .648 & .70 & .399 & 2.78 \\
 & 2 & .821 & .782 & .678 & .75 & .417 & 3.19 \\
 & 3 & .826 & .790 & .689 & .77 & .423 & 3.36 \\
 & 4 & .828 & .789 & .690 & .76 & .419 & 3.29 \\
\bottomrule
\end{tabular}
\end{table*}

%% ============================================================
\subsection{Diagnostic Case Study}
\label{sec:exp-diagnostic}
Because \methodname{} reads the fixed predictor's internal representations rather than its final binary output, it can surface cross-drug interaction evidence even when the predictor misclassifies a pair. We illustrate this with Nitazoxanide + Remdesivir, a pair with literature-reported synergy~\cite{bobrowski2021sarscov2synergy} that the predictor classifies as non-synergistic ($s_{AB} = 0.487$, below the 0.5 classification threshold defined in Section~\ref{sec:problem}).
\begin{figure}[t]
  \centering
  \includegraphics[width=\linewidth]{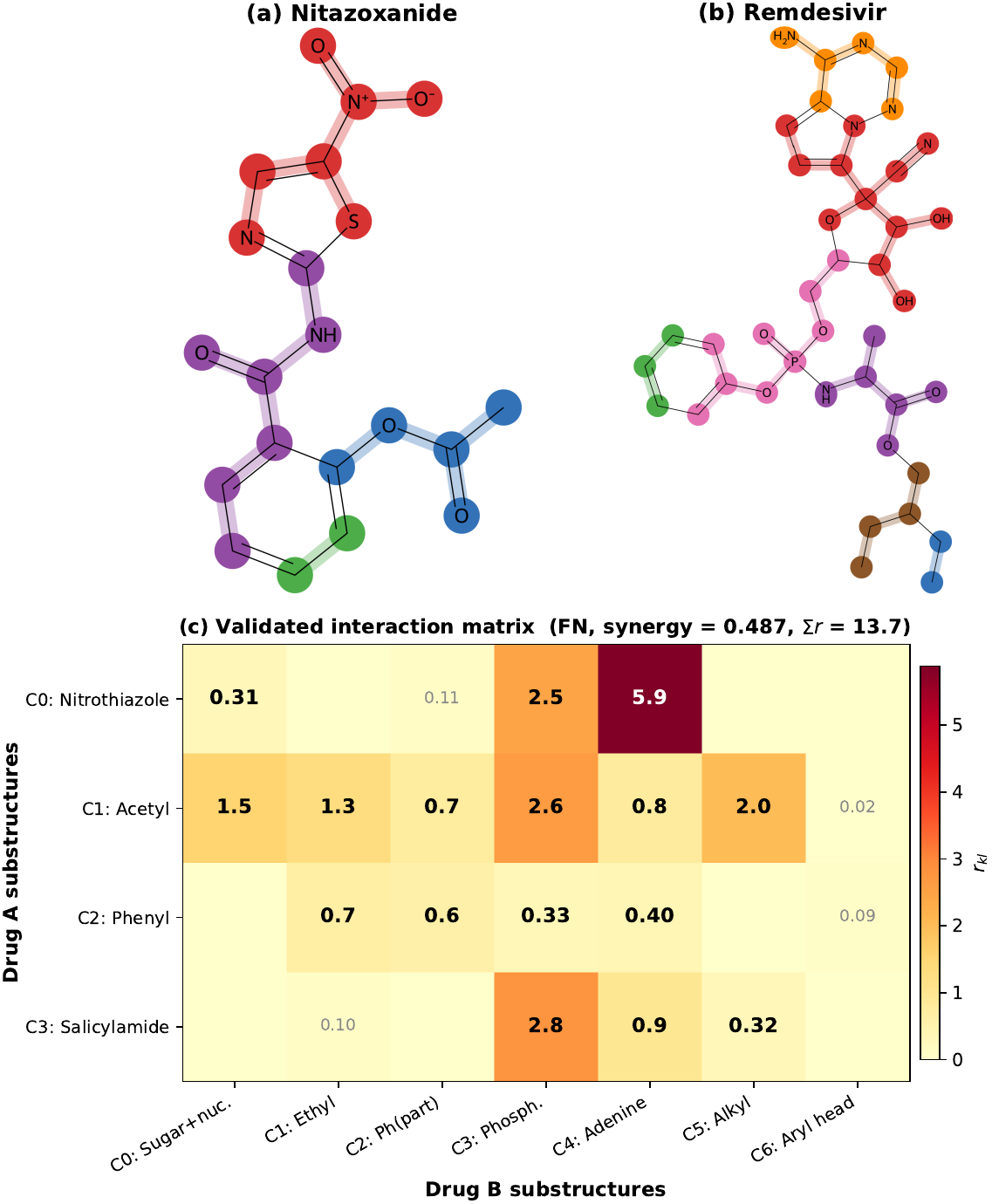}
  \caption{Diagnostic case study: Nitazoxanide + Remdesivir ($s_{AB} = 0.487$, false negative). (a,b)~Motif clusters for each drug. (c)~Validated interaction matrix $R$.}
  \label{fig:fn-diagnostic}
\end{figure}
\textbf{\methodname{} recovers strong interaction evidence in this misclassified pair.} For correctly classified synergistic pairs, the mean top validated interaction score is 4.82; for predictor true negatives it is 1.43. In this false-negative pair, the top score reaches 5.9 (between the nitrothiazole region of Nitazoxanide and the adenine ring of Remdesivir; Figure~\ref{fig:fn-diagnostic}), placing it above the synergistic group mean despite the incorrect negative label. This suggests that synergy-relevant cross-drug evidence remains present in the predictor's internal representations despite not being reflected in the final binary decision. A researcher reviewing \methodname{}'s outputs could flag this pair for further investigation. An additional true-positive case study (Amodiaquine + Nitazoxanide) is provided in Appendix~\ref{app:case-study}.

%% ============================================================
%% §6 DISCUSSION / LIMITATIONS
%% ============================================================
\section{Limitations and Ethical Considerations}
\label{sec:limitations}

\methodname{} is a post-training explanation framework and therefore
inherits the behavior and assumptions of the fixed predictor it explains.
Its region-pair scores characterize predictor-level evidence and should
not be interpreted as direct evidence of biological causality or clinical
efficacy. Our literature-grounded evaluation covers 25 benchmark entries
for which published evidence supports region-level annotation, and its
conclusions remain conditioned on the compounds and assay settings in the
underlying benchmark. Extending evaluation to larger combination screens
and experimentally characterized molecular interactions is an important
direction for future work.

If the predictor has learned spurious correlations, \methodname{}'s
explanation will reflect those correlations rather than true
pharmacological signals; pairing explanation with
predictive uncertainty estimates~\cite{wollschlager2023uncertainty,garain2026uncertainty_calibration}
could help identify such cases. The framework currently requires a
compatible interaction-aware predictor that exposes atom-level
representations and cross-drug signals; applicability to predictors
without these interfaces would require architectural adaptation.

This study, which uses publicly available molecular and drug-combination data, involves no identifiable personal data or human-subject intervention.
\methodname{} is intended to support mechanistic inspection and
experimental prioritization; its outputs should be interpreted together
with domain expertise and experimental evidence rather than used directly
for treatment selection or clinical decision making.

\section{Generative AI Usage}
\label{sec:genai}

Generative AI tools were used to assist with code generation and
debugging, language revision, organization, and literature
and citation cross-checking. All methodological decisions, experimental
design, analyses, reported results, and final claims were determined and
verified by the authors, who take full responsibility for the content of
the paper.

\section{Conclusion}
\label{sec:conclusion}

We set out to answer a specific question: given a fixed drug synergy
predictor, how can we identify which cross-drug motif pairs jointly
contribute to its synergy decision, in a way that is learned and
perturbation-validated rather than predefined or read from a single
forward pass? \methodname{} addresses this through a closed
explanation--validation loop that constructs chemically coherent motifs,
validates retained motif pairs through repeated local perturbations, and
feeds the validated evidence back to refine the motifs. On the 25-pair
literature-annotated subset, \methodname{} achieves a mean motif recall
of 0.826, compared with 0.49--0.66 for baselines. Across the full
71-pair test set, its validated interaction scores yield a TP/TN
separation of 3.36, compared with 0.48--1.49 for the controlled
baselines. These results show that closed-loop perturbation-based
validation improves recovery of literature-supported molecular regions
while producing interaction scores that better reflect the fixed
predictor's behavior.
\FloatBarrier
%% ============================================================
%% ACKNOWLEDGMENTS
%% ============================================================
%\begin{acks}
% TODO: funding acknowledgments
%\end{acks}

%% ============================================================
%% REFERENCES
%% ============================================================
\vspace{-0.5em}
\bibliographystyle{ACM-Reference-Format}
\bibliography{references}

% ============================================================
% APPENDIX — Literature-Grounded Evaluation Protocol Details
% ============================================================
\clearpage
\appendix

%% ============================================================
\section{Literature-Grounded Evaluation Protocol: Detailed Construction}
\label{app:eval-protocol}

This appendix details the construction of the literature-supported
molecular-region reference used in Section~\ref{sec:exp-protocol}.
The reference is designed to evaluate whether an explanation method
localizes pharmacologically relevant regions of the two component drugs
in the SARS-CoV-2 combination benchmark.

Starting from the 71 test entries, we applied a common literature
screening and annotation procedure and retained 25 entries for which
published evidence provided sufficient molecular specificity for
region-level annotation. The resulting reference contains 111
drug-specific molecular-region annotations. The annotations were
constructed independently of explanation outputs and fixed before
method comparison.

We organize the supporting evidence into three provenance tiers.
\textbf{T1} denotes direct experimental combination-response evidence
for the specific drug combination; \textbf{T2} denotes mechanistic
characterization of the component drugs together with an explicit
combination rationale; and \textbf{T3m} denotes mechanism-supported
pharmacological context sufficient to identify molecular regions
relevant to the component drugs. The final set contains 10 T1 entries,
3 T2 entries, and 12 T3m entries.

The construction follows four levels:
\[
\begin{aligned}
\text{published evidence}
&\rightarrow \text{pair-level mechanistic context}\\
&\rightarrow \text{drug-specific molecular region}\\
&\rightarrow \text{atom-index reference set}.
\end{aligned}
\]
This organization separates pair-level pharmacological context from
region-level localization. The literature reference is used for
molecular-region coverage evaluation, while cross-drug region-pair
scores are assessed by the perturbation-based validation procedure
described in the method.

\subsection{Level 1: Literature Evidence and Provenance}
\label{app:sources}

We use published combination studies, antiviral screens, mechanistic
studies, and structure--activity analyses to establish the evidence
chain for each annotated entry. Direct combination-response studies
provide experimental context for specific drug combinations, whereas
drug-level mechanism and structure--activity studies identify
pharmacologically meaningful regions that can be mapped to molecular
structures.

Table~\ref{tab:lit-evidence} lists key primary and mechanistic sources
used to anchor the construction and the representative examples shown
below. Identifiers are reported directly to make the evidence
traceable to the corresponding publications.

\begin{table*}[!htbp]
\centering
\caption{Key literature evidence used in the reference construction.
Combination-response sources establish drug-pair behavior, whereas
mechanistic and structure--activity sources support drug-specific
molecular-region annotation.}
\label{tab:lit-evidence}
\small
\begin{tabular}{p{3.0cm}p{4.3cm}p{7.8cm}}
\toprule
Evidence Role & Identifier & Publication \\
\midrule
Combination response
& PMID:33333292; DOI:10.1016/j.ymthe.2020.12.016
& Bobrowski et al., \emph{Synergistic and Antagonistic Drug Combinations against SARS-CoV-2} (2021). \\

Combination response
& PMID:32251767; DOI:10.1016/j.antiviral.2020.104786
& Choy et al., \emph{Remdesivir, lopinavir, emetine, and homoharringtonine inhibit SARS-CoV-2 replication in vitro} (2020). \\

Combination response
& PMID:34572416; DOI:10.3390/biomedicines9091230
& Kongsomros et al., \emph{Anti-SARS-CoV-2 Activity of Extracellular Vesicle Inhibitors: Screening, Validation, and Combination with Remdesivir} (2021). \\

Combination response
& PMID:36190406; DOI:10.1128/spectrum.03331-22
& Wagoner et al., \emph{Combinations of Host- and Virus-Targeting Antiviral Drugs Confer Synergistic Suppression of SARS-CoV-2} (2022). \\

Drug mechanism
& PMID:33711336; DOI:10.1016/j.antiviral.2021.105056
& Kumar et al., \emph{Emetine suppresses SARS-CoV-2 replication by inhibiting interaction of viral mRNA with eIF4E} (2021). \\

Combination response / mechanism
& PMID:35215969; DOI:10.3390/v14020374
& Sacramento et al., \emph{Unlike Chloroquine, Mefloquine Inhibits SARS-CoV-2 Infection in Physiologically Relevant Cells} (2022). \\

Drug mechanism
& PMID:33465165; DOI:10.1371/journal.ppat.1009212
& Ou et al., \emph{Hydroxychloroquine-mediated inhibition of SARS-CoV-2 entry is attenuated by TMPRSS2} (2021). \\

Drug mechanism
& PMID:33676899; DOI:10.1016/j.ebiom.2021.103255
& Hoffmann et al., \emph{Camostat mesylate inhibits SARS-CoV-2 activation by TMPRSS2-related proteases and its metabolite GBPA exerts antiviral activity} (2021). \\

Structure--activity
& PMID:34898207; DOI:10.1021/acs.jcim.1c01061
& Freidel and Armen, \emph{Modeling the Structure--Activity Relationship of Arbidol Derivatives and Other SARS-CoV-2 Fusion Inhibitors Targeting the S2 Segment of the Spike Protein} (2021). \\

Drug mechanism
& PMID:32284326; DOI:10.1074/jbc.RA120.013679
& Gordon et al., \emph{Remdesivir is a direct-acting antiviral that inhibits RNA-dependent RNA polymerase from severe acute respiratory syndrome coronavirus 2 with high potency} (2020). \\
\bottomrule
\end{tabular}
\end{table*}

\subsection{Level 2: Pair-Level Mechanistic Context}
\label{app:hypotheses}

For each retained benchmark entry, the literature is first summarized
at the drug-pair level. This step records the experimentally observed
combination behavior when direct combination data are available and
relates the pharmacological mechanisms of the two component drugs.
The purpose of this level is to establish the biological context from
which drug-specific regions are subsequently localized.

Table~\ref{tab:app-hypotheses} gives representative examples. The
examples span direct combination evidence and complementary
mechanistic evidence, illustrating how the same annotation procedure
is applied across the evidence hierarchy.

\begin{table*}[!htbp]
\centering
\caption{Representative pair-level mechanistic context used to guide
drug-specific region annotation.}
\label{tab:app-hypotheses}
\small
\begin{tabular}{p{1.8cm}p{3.6cm}p{10.0cm}}
\toprule
Pair ID & Drugs & Literature-Grounded Context \\
\midrule

pair\_000
& Amodiaquine + Nitazoxanide
& Experimental combination screening reports synergy for nitazoxanide
with amodiaquine. The two drugs contribute distinct antiviral
pharmacological scaffolds, motivating localization of the
chloroquinoline/basic-amine regions of amodiaquine and the thiazolide
regions of nitazoxanide. \\

pair\_002
& Nitazoxanide + Remdesivir
& Direct combination screening reports significant synergy between
nitazoxanide and remdesivir. Nitazoxanide contributes a thiazolide
antiviral scaffold, whereas remdesivir acts through its
nucleotide-analog/RdRp axis. These mechanisms define complementary
drug-specific regions for localization analysis. \\

pair\_023
& Emetine + Remdesivir
& In vitro experiments report synergistic inhibition of SARS-CoV-2 by
emetine and remdesivir. Emetine suppresses viral protein synthesis and
has been linked to inhibition of viral mRNA interaction with eIF4E,
whereas remdesivir directly inhibits viral RdRp. The two mechanisms
provide distinct post-entry pharmacological contexts for molecular
region annotation. \\

pair\_061
& Camostat + Remdesivir
& Camostat inhibits SARS-CoV-2 entry through TMPRSS2-related serine
proteases, whereas remdesivir targets viral RNA synthesis through
RdRp. The entry-versus-replication distinction provides a
mechanistically complementary basis for localizing relevant regions
on the two component drugs. \\

pair\_067
& Mefloquine + Remdesivir
& Mefloquine has been shown to reduce SARS-CoV-2 entry in
physiologically relevant cells and to enhance the antiviral activity
of remdesivir. Remdesivir provides the complementary
nucleotide-analog/RdRp mechanism, motivating region annotations on
both drug structures. \\

\bottomrule
\end{tabular}
\end{table*}

\subsection{Level 3: Drug-Specific Literature-Supported Regions}
\label{app:loci}

Pair-level mechanistic context is next resolved into drug-specific
molecular regions. A region corresponds to a chemically coherent
functional group, scaffold component, or pharmacophore-level locus
that can be associated with the published pharmacology of the drug.
This provides a common structural unit against which explanation
methods can be evaluated.

The annotation is intentionally region-based rather than restricted to
individual atoms: many pharmacological determinants span aromatic
systems, basic side chains, modified nucleosides, or prodrug moieties.
Table~\ref{tab:app-loci} shows representative region definitions.

\begin{table*}[!htbp]
\centering
\caption{Representative drug-specific literature-supported molecular
regions used in the localization evaluation.}
\label{tab:app-loci}
\small
\begin{tabular}{p{1.8cm}p{0.8cm}p{4.2cm}p{4.0cm}p{5.1cm}}
\toprule
Pair ID & Drug & Region & Chemical Type & Annotation Role \\
\midrule

pair\_000
& A
& Chloroquinoline core
& Heterocyclic aromatic
& Principal heteroaromatic scaffold of amodiaquine \\

pair\_000
& A
& Diethylamino sidechain
& Basic amine
& Basic side-chain region associated with the drug's physicochemical and lysosomotropic behavior \\

pair\_000
& B
& Nitrothiazole region
& Heteroaromatic motif
& Characteristic thiazolide heteroaromatic region of nitazoxanide \\

pair\_000
& B
& Salicylamide core
& Aromatic amide
& Aromatic salicylamide scaffold region \\

\midrule

pair\_002
& A
& Nitrothiazole region
& Heteroaromatic motif
& Characteristic thiazolide heteroaromatic region of nitazoxanide \\

pair\_002
& A
& Salicylamide core
& Aromatic amide
& Aromatic salicylamide scaffold region \\

pair\_002
& B
& Adenine-like nucleobase
& Nucleobase
& Nucleobase-recognition region of the remdesivir nucleotide analog \\

pair\_002
& B
& Sugar--nitrile core
& Modified ribose
& Modified ribose region contributing to the nucleotide-analog scaffold \\

pair\_002
& B
& Phosphoramidate aryl head
& Prodrug motif
& ProTide region associated with intracellular formation of the active nucleotide analog \\

\midrule

pair\_061
& A
& Guanidinium benzoate region
& Serine-protease inhibitor motif
& Recognition region associated with camostat's serine-protease inhibitory pharmacology \\

pair\_061
& A
& Dimethylcarbamoyl ester tail
& Ester-containing region
& Peripheral ester-containing region of the camostat scaffold \\

pair\_061
& B
& Adenine-like nucleobase
& Nucleobase
& Nucleobase-recognition region of the remdesivir nucleotide analog \\

pair\_061
& B
& Phosphoramidate aryl head
& Prodrug motif
& ProTide region associated with intracellular activation \\

\bottomrule
\end{tabular}
\end{table*}

\subsection{Level 4: Atom-Index Reference Mapping}
\label{app:atom-mapping}

Each literature-supported molecular region is finally converted to an
atom-index reference set on the canonicalized molecular graph.
Indices follow zero-based RDKit atom indexing. The mapping preserves
the functional-group or scaffold boundary represented by the region;
numerical indices therefore need not be consecutive when a chemically
coherent region spans multiple branches of the molecular graph.

For a reference region $L$ and a predicted region $P$, coverage is
evaluated through their atom-set overlap. Table~\ref{tab:app-atoms}
shows representative mappings together with the corresponding
region-aware matching results.

\subsection{Region-Aware Matching Protocol}
\label{app:region-matching}

Because learned cluster boundaries need not coincide exactly with
literature-defined functional-group boundaries, we use a constrained
region-aware matching rule. For each reference region $L$, candidate
predictions are restricted to regions from the same benchmark entry
and the same drug. We consider either a single predicted cluster or a
union of at most two clusters. For a candidate union $P$, its size is
restricted by

\begin{equation}
|P|
\leq
\max
\left(
|L|+3,\;
\left\lceil 1.4|L| \right\rceil
\right).
\end{equation}

Among admissible candidates, the best matching region is selected
under the same rule for all compared methods. Region recall,
precision, and Jaccard overlap are

\begin{equation}
\mathrm{Recall}(L,P)
=
\frac{|L\cap P|}{|L|},
\qquad
\mathrm{Precision}(L,P)
=
\frac{|L\cap P|}{|P|},
\end{equation}

\begin{equation}
\mathrm{Jaccard}(L,P)
=
\frac{|L\cap P|}{|L\cup P|}.
\end{equation}

We additionally report the fraction of reference regions reaching
recall $\geq 0.7$. The same reference atom sets and matching
constraints are used for every explanation method.

\begin{table*}[!htbp]
\centering
\caption{Representative atom-level reference mappings and
region-aware matching results. Atom indices are zero-based RDKit
indices. $|\mathrm{ref}|$ and $|\mathrm{pred}|$ denote the reference
and matched predicted region sizes, respectively.}
\label{tab:app-atoms}
\small
\begin{tabular}{lllllcccc}
\toprule
Pair & Drug & Region & Atom Indices &
$|\mathrm{ref}|$ & $|\mathrm{pred}|$ &
Recall & Precision & Jaccard \\
\midrule

pair\_000 & A & Chloroquinoline core
& \{10--20\} & 11 & 11 & 1.000 & 1.000 & 1.000 \\

pair\_000 & A & Diethylamino sidechain
& \{0--5\} & 6 & 8 & 1.000 & 0.750 & 0.750 \\

pair\_000 & B & Nitrothiazole region
& \{13--20\} & 8 & 8 & 0.875 & 0.875 & 0.778 \\

pair\_000 & B & Salicylamide core
& \{4--12\} & 9 & 13 & 0.889 & 0.615 & 0.571 \\

\midrule

pair\_002 & A & Nitrothiazole region
& \{13--20\} & 8 & 7 & 0.875 & 1.000 & 0.875 \\

pair\_002 & A & Salicylamide core
& \{4--12\} & 9 & 9 & 0.889 & 0.889 & 0.800 \\

pair\_002 & B & Adenine-like nucleobase
& \{21--30\} & 10 & 7 & 0.600 & 0.857 & 0.545 \\

pair\_002 & B & Phosphoramidate aryl head
& \{11--15,35--41\} & 12 & 12 & 0.917 & 0.917 & 0.846 \\

pair\_002 & B & Sugar--nitrile core
& \{16--20,31--34\} & 9 & 12 & 0.889 & 0.667 & 0.615 \\

\midrule

pair\_023 & A & Amine polycyclic core
& \{2--6,17--24\} & 13 & 18 & 0.923 & 0.667 & 0.632 \\

pair\_023 & A & Dimethoxy aromatic face A
& \{7--16\} & 10 & 10 & 0.900 & 0.900 & 0.818 \\

pair\_023 & A & Dimethoxy aromatic face B
& \{25--34\} & 10 & 13 & 1.000 & 0.769 & 0.769 \\

pair\_023 & B & Adenine-like nucleobase
& \{21--30\} & 10 & 9 & 0.900 & 1.000 & 0.900 \\

pair\_023 & B & Phosphoramidate aryl head
& \{11--15,35--41\} & 12 & 16 & 0.917 & 0.688 & 0.647 \\

\midrule

pair\_061 & A & Guanidinium benzoate region
& \{14--26\} & 13 & 17 & 1.000 & 0.765 & 0.765 \\

pair\_061 & A & Dimethylcarbamoyl ester tail
& \{0--9\} & 10 & 12 & 0.900 & 0.750 & 0.692 \\

pair\_061 & B & Adenine-like nucleobase
& \{21--30\} & 10 & 14 & 1.000 & 0.714 & 0.714 \\

pair\_061 & B & Phosphoramidate aryl head
& \{11--15,35--41\} & 12 & 16 & 1.000 & 0.750 & 0.750 \\

\bottomrule
\end{tabular}
\end{table*}

\subsection{Reference-Set Statistics}
\label{app:summary}

Table~\ref{tab:app-summary} summarizes the final reference used for
literature-region coverage evaluation. Across the 25 annotated
benchmark entries, the reference contains 111 molecular regions, or
4.44 regions per entry on average.

\begin{table}[!htbp]
\centering
\caption{Summary statistics of the literature-supported
molecular-region reference.}
\label{tab:app-summary}
\small
\begin{tabular}{lr}
\toprule
Statistic & Value \\
\midrule
Annotated benchmark entries & 25 \\
Literature-supported reference regions & 111 \\
Mean reference regions per entry & 4.44 \\
Reference regions on Drug A & 57 \\
Reference regions on Drug B & 54 \\
T1 entries: direct combination-response evidence & 10 \\
T2 entries: mechanism + combination rationale & 3 \\
T3m entries: mechanism-supported context & 12 \\
\bottomrule
\end{tabular}
\end{table}

\FloatBarrier

%% ============================================================
\section{Case Study}
\label{app:case-study}

We provide a true-positive case study to illustrate how \methodname{}
connects molecular segmentation, cross-drug interaction scoring, and
literature-supported region localization.

\begin{figure*}[t]
  \centering
  \includegraphics[width=\textwidth]{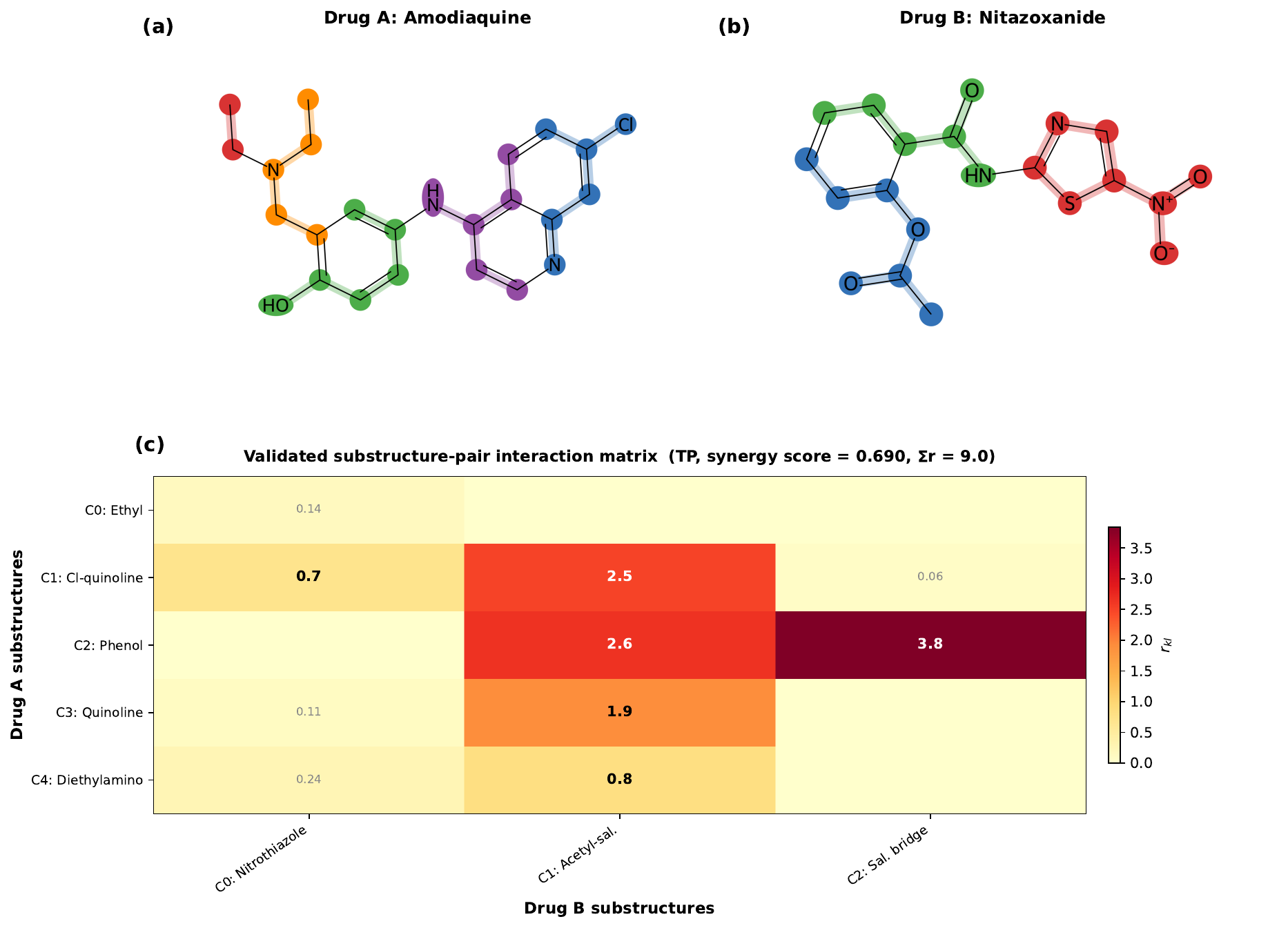}
  \caption{Case study of Amodiaquine + Nitazoxanide
  (true positive; predicted synergy score 0.690).
  (a)~Predicted substructure segmentation of Amodiaquine.
  (b)~Predicted substructure segmentation of Nitazoxanide.
  (c)~Region-pair interaction matrix $R$ after local perturbation
  validation, where $r_{kl}$ denotes the validated interaction score
  between predicted regions $k$ and $l$.}
  \label{fig:case-pair000}
\end{figure*}

Figure~\ref{fig:case-pair000} shows Amodiaquine + Nitazoxanide, a
true-positive pair with a predicted synergy score of 0.690.

Drug A (Amodiaquine) is segmented into five predicted substructures
(panel~a). These include an ethyl-containing region (C0), a
chlorinated-quinoline region (C1), a phenol-containing region (C2), a
quinoline region (C3), and the diethylamino sidechain (C4). In
particular, C1 and C3 together cover the broader
chloroquinoline-containing scaffold, while C4 captures the
diethylamino sidechain.

Drug B (Nitazoxanide) is segmented into three predicted substructures
(panel~b): the nitrothiazole-containing region (C0), an
acetyl-salicylate-containing region (C1), and the salicylamide-bridge
region (C2). The resulting segmentation separates the major
heteroaromatic and aromatic-amide components of the molecule into
distinct regions for cross-drug interaction analysis.

The region-pair interaction matrix $R$ (panel~c) shows that the
strongest effect after local perturbation validation occurs between
the phenol-containing region of Amodiaquine (C2) and the
corresponding \mbox{salicylamide-bridge} region of Nitazoxanide (C2), with
$r_{kl}=3.8$. Other prominent interactions are also concentrated
among a small number of region pairs, including C2$\times$C1
($r_{kl}=2.6$), C1$\times$C1 ($r_{kl}=2.5$; chlorinated quinoline),
and C3$\times$C1 ($r_{kl}=1.9$; quinoline). Together, these scores
show that the validated interaction signal is concentrated on a
small subset of cross-drug region pairs rather than being distributed
uniformly across the two molecules.

The high-scoring interaction pattern is also consistent with the
literature-region evaluation. On Amodiaquine, the predicted
segmentation recovers the literature-supported chloroquinoline core
and diethylamino sidechain. On Nitazoxanide, the predicted regions
overlap the annotated nitrothiazole and salicylamide regions. Across
the four literature-supported reference regions for this benchmark
entry, \methodname{} achieves a mean recall of 0.941, including
recall of 1.0 for both the chloroquinoline core and the
diethylamino sidechain. This example illustrates how
\methodname{} jointly provides chemically localized molecular regions
and cross-drug interaction scores supported by local perturbation
validation.

%% ============================================================
\section{External Baseline Adaptation}
\label{app:baseline-setup}

All external baselines explain the same fixed predictor. Because these methods are designed for single-graph predictions, each drug in a pair is explained separately. Atom-level importance scores are converted into candidate regions via two strategies: connected components of top-$k$ atoms, and spectral clustering on the importance-weighted induced subgraph. For each baseline, $k$ is swept globally and the configuration with highest mean recall is reported.

PGExplainer is trained on all available drug molecules. SubgraphX returns ranked connected subgraphs; the top-3 are used for union selection. CF-GNNExplainer sweeps sparsity threshold (0.3--0.8) and regularization weight ($\lambda \in \{0.01, 0.05, 0.1, 0.5\}$). The random substructure baseline samples size-matched connected subgraphs via BFS (200 trials per molecule). All baselines use the same region-aware matching protocol as \methodname{}: a reference motif may match a single predicted region or a union of at most two adjacent regions sharing at least one bond, subject to the same size cap.

%% ============================================================
\section{Method Implementation Details}
\label{app:method-details}

This appendix provides the exact definitions of the operators abstracted in the main text. All notation follows Section~\ref{sec:method}. The constructions below are described for drug~$A$; drug~$B$ is treated identically.

\subsection{Structural Affinity}

The structural view encodes molecular locality via a Gaussian decay on shortest-path graph distance:
\begin{equation}
  W_A^{\mathrm{struct}}[i,j] = \mathbf{1}[d_A(i,j) \le d_s] \cdot \exp\!\left(-\frac{d_A(i,j)^2}{2\sigma_s^2}\right),
\end{equation}
where $d_A(i,j)$ is the shortest-path distance between atoms $i$ and $j$ in the molecular graph of drug~$A$, $d_s$ controls the neighborhood radius, and $\sigma_s$ controls the decay rate.

\subsection{Interaction-Pattern Affinity}

The interaction-pattern view connects nearby atoms whose cross-drug evidence profiles are both sufficiently active and similar. We first define, for each atom $i$, its cross-drug evidence magnitude and normalized profile:
\begin{equation}
  a_i = \|M[i,:]\|_2, \qquad p_i = \frac{M[i,:]}{\max(\|M[i,:]\|_2,\, \epsilon)}.
\end{equation}
An activity gate suppresses misleading similarities between nearly inactive profiles:
\begin{equation}
  m(a_i) = \sigma\!\bigl(\beta\,(a_i - \tau_{\mathrm{act}})\bigr),
\end{equation}
where $\sigma$ is the logistic sigmoid, $\beta$ controls gate sharpness, and $\tau_{\mathrm{act}}$ is the activity threshold (set to the 60th percentile of $\{a_i\}$). The interaction-pattern affinity is:
\begin{equation}
  W_A^{\mathrm{pattern}}[i,j] = \mathbf{1}[d_A(i,j) \le d_p] \cdot \exp\!\left(-\frac{d_A(i,j)^2}{2\sigma_p^2}\right) \cdot m(a_i)\,m(a_j) \cdot p_i^\top p_j.
\end{equation}
Profile similarity captures whether two atoms interact with the partner drug in a similar way, while the activity gate ensures that this similarity is meaningful rather than driven by near-zero profiles.

\subsection{Normalized Laplacian}

Before motif assignment, each affinity matrix $W$ is converted into a self-looped symmetric normalized Laplacian:
\begin{equation}
  \widehat{W} = W + I, \qquad L_{\mathrm{norm}} = I - \widehat{D}^{-1/2}\,\widehat{W}\,\widehat{D}^{-1/2},
\end{equation}
where $\widehat{D}$ is the diagonal degree matrix of $\widehat{W}$.

\subsection{Assignment Regularization}

The regularizer $\mathcal{R}(S)$ in Eq.~\eqref{eq:assignment-loss} combines two terms:
\begin{equation}
  \mathcal{R}(S) = -\lambda_H \,\mathcal{R}_{\mathrm{ent}}(S) + \rho_{\mathrm{mass}}\,\mathcal{R}_{\mathrm{mass}}(S).
\end{equation}
The entropy term $\mathcal{R}_{\mathrm{ent}}(S) = -\sum_{i,k} S[i,k]\log(S[i,k]+\epsilon)$ prevents premature collapse of the soft assignment into hard one-hot vectors. The minimum-mass term $\mathcal{R}_{\mathrm{mass}}(S) = \sum_k \mathrm{softplus}(m_{\min} - \sum_i S[i,k])$ penalizes clusters whose total soft mass falls below a threshold $m_{\min}$, discouraging pathologically small motifs.

\subsection{Ring Completion}

After hard assignment $\hat{k}(i) = \arg\max_k S[i,k]$, graph connectivity is enforced so that every resulting motif induces a connected subgraph of the molecular graph. Partially assigned ring systems are then completed: if a motif already contains at least half the atoms of a ring (identified via RDKit \texttt{GetRingInfo}), the remaining ring atoms are absorbed into that motif. This prevents chemically indivisible ring systems from being split across motifs.

\subsection{Feedback Operator}

For each atom $i$ in drug~$A$, the validated interaction profile summarizes its connection to drug~$B$'s motifs through the current validated scores:
\begin{equation}
  v_A^{(t)}(i,:) = \sum_k S_A^{(t)}[i,k] \cdot R^{(t)}[k,:].
\end{equation}
The feedback affinity between atoms $i$ and $j$ combines graph locality with safe cosine similarity of their validated profiles:
\begin{equation}
  \widetilde{W}_A^{\mathrm{fb},(t+1)}[i,j] = \mathbf{1}[d_A(i,j) \le d_{\mathrm{fb}}] \cdot \exp\!\left(-\frac{d_A(i,j)^2}{2\sigma_{\mathrm{fb}}^2}\right) \cdot \frac{v_i^\top v_j}{\max(\|v_i\|_2\,\|v_j\|_2,\,\epsilon)}.
\end{equation}
The feedback view is then smoothed via exponential moving average:
\begin{equation}
  W_A^{\mathrm{fb},(t+1)} = (1 - \alpha_{\mathrm{fb}})\,W_A^{\mathrm{fb},(t)} + \alpha_{\mathrm{fb}}\,\widetilde{W}_A^{\mathrm{fb},(t+1)}.
\end{equation}

\subsection{Complete Hyperparameter Table}

Table~\ref{tab:full_hyperparams} lists all hyperparameters of the explanation framework.

\begin{table*}[!htbp]
\centering
\caption{Complete hyperparameter settings for the explanation framework.}
\label{tab:full_hyperparams}
\small
\begin{tabular}{lll}
\toprule
Parameter & Default & Role \\
\midrule
Target motif size $c$ & 6 & Atoms per motif (determines $K_d = \mathrm{clip}(\lfloor N_d/c \rceil, 2, 8)$) \\
Perturbation trials $T$ & 16 & Trials per retained motif pair \\
Outer-loop iterations & 3 & Default refinement depth \\
Candidate screening $\theta_{\mathrm{screen}}$ & top 30\%, min 3, max 20 & Pair-selection scope \\
$(\lambda_{\mathrm{struct}}, \lambda_{\mathrm{pattern}}, \lambda_{\mathrm{fb}})$ & $(1.0,\; 0.7,\; 0.3)$ & Affinity-view weights \\
$\lambda_H$ & 0.03 & Entropy regularization weight \\
$\rho_{\mathrm{mass}}$ & 0.10 & Minimum-mass penalty weight \\
$\alpha_{\mathrm{fb}}$ & 0.5 & Feedback EMA rate \\
$\tau_{\mathrm{act}}$ & 60th percentile & Interaction-pattern activity gate \\
$(d_s, \sigma_s)$ & $(5.0,\; 1.5)$ & Structural affinity parameters \\
$(d_p, \sigma_p)$ & $(4.0,\; 1.25)$ & Pattern affinity parameters \\
$(d_{\mathrm{fb}}, \sigma_{\mathrm{fb}})$ & $(5.0,\; 2.0)$ & Feedback affinity parameters \\
\bottomrule
\end{tabular}
\end{table*}

%% ============================================================
\section{Perturbation-Based Validation Details}
\label{app:perturbation-details}

This appendix provides the exact perturbation procedure abstracted in Section~\ref{sec:phase4}.

\subsection{Local Subset Sampling}

For each retained motif pair $(k,l)$ and trial $t$, we independently sample local subsets of atoms from the two motifs:
\begin{equation}
  \mathcal{M}_{A,k}^{(t)} \subseteq \mathcal{G}_A^{(k)}, \qquad \mathcal{M}_{B,l}^{(t)} \subseteq \mathcal{G}_B^{(l)},
\end{equation}
where $\mathcal{G}_A^{(k)}$ denotes the set of atoms assigned to motif $k$ in drug~$A$. Each trial uses a different subset and masking fraction, so the $T{=}16$ trials produce a distribution of local perturbations for the same motif pair rather than repeating an identical intervention.

\subsection{Feature Substitution}

For each selected atom $i \in \mathcal{M}_{A,k}^{(t)}$, the atom-level representation is replaced by a learned neutral mask embedding:
\begin{equation}
  H_A^{\mathrm{pert}}[i,:] = e_{\mathrm{mask}}, \qquad e_{\mathrm{mask}} \in \mathbb{R}^{d_h}.
\end{equation}
Drug~$B$ is treated identically. No nodes or edges are removed: the molecular topology remains unchanged throughout.

\subsection{Local Reconditioning}

After feature substitution, the representations in the 2-hop neighborhood of the masked motif are locally reconditioned:
\begin{equation}
  \mathcal{N}_A^{(k)} = \{i : d_A(i, \mathcal{G}_A^{(k)}) \le 2\}.
\end{equation}
A 2-layer local message-passing network operates on $\mathcal{N}_A^{(k)}$ to allow nearby atom representations to adjust to the masked features, producing a locally consistent perturbed state without propagating the perturbation signal through the entire molecular graph. Its parameters are learned only during the one-time mask-aware calibration and remain frozen during explanation.

\subsection{Perturbation States and Interaction Effect}

For each trial $t$, four perturbation configurations yield four predictor outputs $s_{11}^{(t)}, s_{10}^{(t)}, s_{01}^{(t)}, s_{00}^{(t)}$ (both regions retained, only drug~$A$'s region retained, only drug~$B$'s region retained, both masked). The perturbation-derived interaction effect is the second-order finite difference given in Eq.~\eqref{eq:interaction-effect}.

\subsection{Stable Interaction Score}

Across $T$ trials, the effect distribution is summarized by:
\begin{align}
  \mu_{kl} &= \frac{1}{T}\sum_t I_{kl}^{(t)}, \qquad \sigma_{kl} = \mathrm{std}(\{I_{kl}^{(t)}\}_t), \\
  p_{kl}^{+} &= \frac{1}{T}\sum_t \mathbf{1}[I_{kl}^{(t)} > 0], \qquad q_{kl} = \frac{1}{T}\sum_t \mathbf{1}[|I_{kl}^{(t)}| > \tau_I].
\end{align}
The validated interaction score combines these four statistics:
\begin{equation}
  r_{kl} = \mathrm{softplus}\!\left(\frac{\mu_{kl}}{\sigma_{kl}+\epsilon}\right) \cdot \max(0,\, 2p_{kl}^{+}-1) \cdot q_{kl}.
\end{equation}

%% ============================================================
\section{Hyperparameter Sensitivity Analysis}
\label{app:sensitivity}

This appendix reports the full sensitivity analysis summarized in Section~\ref{sec:exp-ablation}. Each hyperparameter is varied individually with all others held at their defaults (Table~\ref{tab:full_hyperparams}). We report the same six evaluation metrics used throughout the main text.

\begin{table*}[!htbp]
\centering
\caption{Complete sensitivity results across five hyperparameter sweeps and an early outer-loop convergence analysis. Bold rows indicate defaults for the five swept hyperparameters and the preselected outer-loop depth of 3. The iteration-10 row is included only to characterize convergence beyond the default.}
\label{tab:sensitivity_full}
\small
\begin{tabular}{llcccccc}
\toprule
Parameter & Setting & Recall & Precision & Jaccard & HR${\geq}$0.7 & $\rho$ & TP/TN \\
\midrule
\multirow{5}{*}{$K_d$ ($c$)}
 & 4  & 0.791 & 0.766 & 0.641 & 0.69 & 0.404 & 3.05 \\
 & 5  & 0.823 & 0.788 & 0.686 & 0.75 & 0.418 & 3.28 \\
 & \textbf{6}  & \textbf{0.826} & \textbf{0.790} & \textbf{0.689} & \textbf{0.77} & \textbf{0.423} & \textbf{3.36} \\
 & 8  & 0.802 & 0.741 & 0.632 & 0.71 & 0.409 & 3.14 \\
 & 10 & 0.768 & 0.699 & 0.582 & 0.63 & 0.392 & 2.88 \\
\midrule
\multirow{4}{*}{$T$}
 & 4  & 0.812 & 0.777 & 0.668 & 0.73 & 0.331 & 2.38 \\
 & 8  & 0.821 & 0.785 & 0.681 & 0.75 & 0.401 & 3.02 \\
 & \textbf{16} & \textbf{0.826} & \textbf{0.790} & \textbf{0.689} & \textbf{0.77} & \textbf{0.423} & \textbf{3.36} \\
 & 32 & 0.824 & 0.791 & 0.690 & 0.77 & 0.431 & 3.44 \\
\midrule
\multirow{5}{*}{Outer-loop state}
 & 0  & 0.724 & 0.681 & 0.568 & 0.55 & 0.352 & 1.95 \\
 & 1  & 0.796 & 0.758 & 0.648 & 0.70 & 0.399 & 2.78 \\
 & 2  & 0.821 & 0.782 & 0.678 & 0.75 & 0.417 & 3.19 \\
 & \textbf{3}  & \textbf{0.826} & \textbf{0.790} & \textbf{0.689} & \textbf{0.77} & \textbf{0.423} & \textbf{3.36} \\
 & 4  & 0.828 & 0.789 & 0.690 & 0.76 & 0.419 & 3.29 \\
 & 10  & 0.823 & 0.787 & 0.681 & 0.75 & 0.418 & 3.20 \\
\midrule
\multirow{5}{*}{$\theta_{\mathrm{screen}}$}
 & top 10\% & 0.815 & 0.781 & 0.669 & 0.73 & 0.361 & 2.71 \\
 & top 20\% & 0.823 & 0.787 & 0.684 & 0.75 & 0.414 & 3.24 \\
 & \textbf{top 30\%} & \textbf{0.826} & \textbf{0.790} & \textbf{0.689} & \textbf{0.77} & \textbf{0.423} & \textbf{3.36} \\
 & top 50\% & 0.827 & 0.788 & 0.689 & 0.76 & 0.425 & 3.31 \\
 & top 100\% & 0.822 & 0.782 & 0.683 & 0.75 & 0.415 & 3.22 \\
\midrule
\multirow{4}{*}{$\lambda_{\mathrm{pattern}}$}
 & 0.3 & 0.799 & 0.770 & 0.648 & 0.70 & 0.394 & 2.98 \\
 & 0.5 & 0.824 & 0.789 & 0.687 & 0.76 & 0.419 & 3.30 \\
 & \textbf{0.7} & \textbf{0.826} & \textbf{0.790} & \textbf{0.689} & \textbf{0.77} & \textbf{0.423} & \textbf{3.36} \\
 & 1.0 & 0.814 & 0.758 & 0.652 & 0.71 & 0.421 & 3.29 \\
\midrule
\multirow{5}{*}{$\lambda_{\mathrm{fb}}$}
 & 0.0 & 0.726 & 0.683 & 0.570 & 0.55 & 0.354 & 1.97 \\
 & 0.1 & 0.789 & 0.751 & 0.641 & 0.69 & 0.397 & 2.81 \\
 & \textbf{0.3} & \textbf{0.826} & \textbf{0.790} & \textbf{0.689} & \textbf{0.77} & \textbf{0.423} & \textbf{3.36} \\
 & 0.5 & 0.827 & 0.786 & 0.686 & 0.76 & 0.424 & 3.33 \\
 & 0.7 & 0.811 & 0.765 & 0.655 & 0.72 & 0.412 & 3.14 \\
\bottomrule
\end{tabular}
\end{table*}

Across the five hyperparameter sweeps and the early outer-loop convergence analysis, the framework shows graceful degradation rather than cliff-edge sensitivity. The strongest effects come from feedback-related parameters ($\lambda_{\mathrm{fb}}$ and the outer-loop depth): disabling feedback ($\lambda_{\mathrm{fb}}=0$) reduces recall by approximately 0.10 and TP/TN separation by 1.4, consistent with the ablation findings in Section~\ref{sec:exp-ablation}. Segmentation granularity has a moderate effect, with cluster sizes from 5 to 8~atoms yielding recall above 0.80. Perturbation trials primarily affect discrimination metrics rather than coverage: recall varies by only 0.014 across $T \in \{4,8,16,32\}$, while TP/TN separation ranges from 2.38 to 3.44. The screening threshold has minimal impact on motif coverage: validating all candidate pairs (top~100\%) yields nearly identical recall to the default (top~30\%). The interaction-pattern weight $\lambda_{\mathrm{pattern}}$ shows a mild optimum at 0.7; setting it to 1.0 slightly reduces precision, suggesting that over-weighting interaction-pattern similarity can override the structural prior.

%% ============================================================
\section{Evaluation Robustness Analysis}
\label{app:robustness}

This appendix reports evidence-tier stratification and pair-selection stability analyses summarized in Section~\ref{sec:exp-protocol}.

\paragraph{Evidence-tier stratification.}
Table~\ref{tab:tier_breakdown} reports motif coverage metrics stratified by the evidence tier assigned during reference-set construction (Section~\ref{sec:exp-protocol} and Appendix~\ref{app:eval-protocol}). Tier-1 pairs, supported by direct experimental synergy evidence, achieve the highest recall (0.842) and hit rate (85.8\%), consistent with the expectation that pharmacophore annotations derived from primary screening data align most closely with the predictor's learned motif boundaries. Tier-2 pairs show slightly lower recall (0.809) but higher precision (0.830), reflecting tighter but less complete motif recovery. Tier-3m pairs achieve recall of 0.824, with the widest variance, as expected given that their annotations are derived from mechanism-level rationale rather than direct combination evidence.

\begin{table}[t]
\centering
\caption{Motif coverage stratified by evidence tier. 95\% CIs
from pair-level bootstrap (2,000 resamples within each tier).
The ``All'' row reports pair-level means; the motif-level mean
in Section~5.3 (0.826) weights each motif equally regardless
of pair.}
\label{tab:tier_coverage}

\footnotesize
\setlength{\tabcolsep}{4pt}
\begin{tabular}{lcccc}
\toprule
Tier & \#Pairs & Recall & Precision & HR$\geq$0.7 \\
\midrule
T1  & 10 & 0.842 [0.767--0.898] & 0.814 [0.749--0.868] & 0.858 [0.738--0.958] \\
T2  & 3  & 0.809 [0.753--0.838] & 0.830 [0.709--0.906] & 0.700 [0.500--0.800] \\
T3m & 12 & 0.824 [0.757--0.890] & 0.768 [0.709--0.829] & 0.718 [0.546--0.875] \\
\midrule
All & 25 & 0.829 [0.785--0.871] & 0.794 [0.751--0.835] & 0.771 [0.672--0.861] \\
\bottomrule
\end{tabular}
\end{table}

\paragraph{Pair-selection stability.}
To verify that the aggregate results are not driven by a small number of favorable pairs, we perform leave-$k$-out analysis for $k \in \{1,2,3,5\}$. For each $k$, we randomly remove $k$ pairs from the 25-pair set and recompute mean recall over the remaining pairs, repeating 2{,}000 times. Table~\ref{tab:leave_k_out} reports the resulting stability ranges. Even at $k{=}5$ (removing 20\% of pairs), the 95\% range of mean recall is 0.810--0.852, indicating that no small subset dominates the aggregate. The most influential single pair is pair\_003 (Nitazoxanide + Remdesivir, recall 0.553): removing it raises the mean by +0.012; the most favorable pair is pair\_061 (Camostat + Remdesivir, recall 0.975): removing it lowers the mean by $-$0.006. No individual pair shifts the mean by more than 1.5 percentage points.

\begin{table}[!htbp]
\centering
\caption{Pair-selection stability of mean recall under leave-$k$-out resampling (2{,}000 random subsets per $k$).}
\label{tab:leave_k_out}
\small
\begin{tabular}{ccc}
\toprule
$k$ removed & Pairs remaining & 95\% range of mean recall \\
\midrule
1 & 24 & [0.823--0.841] \\
2 & 23 & [0.819--0.843] \\
3 & 22 & [0.815--0.847] \\
5 & 20 & [0.809--0.852] \\
\bottomrule
\end{tabular}
\end{table}

%% ============================================================
\section{Predictor Implementation and Adequacy}
\label{app:predictor}

\subsection{Architecture}

The predictor combines two molecular encoding branches. A 2D branch based on a directed message-passing neural network (D-MPNN)~\cite{yang2019chemprop} operates on the molecular graph and produces atom-level representations from bond-level messages. A 3D branch based on an equivariant graph neural network (EGNN) operates on a molecular conformer and provides geometry-aware atom representations. The 2D and 3D representations are fused before entering the interaction module.

A bidirectional atom-level cross-attention module then conditions each drug's representation on its partner, producing pair-conditioned atom representations and exposing the cross-drug association matrix $\hat{A} \in \mathbb{R}^{N_A \times N_B}$ used by \methodname{}. Separate prediction heads produce the single-drug activities $P_A$, $P_B$ and the combination activity $P_{AB}$; the synergy score is $s_{AB} = P_{AB} - P_{\mathrm{bliss}}$ (Eq.~\eqref{eq:synergy-score}), and a pair is classified as synergistic when $s_{AB} > 0.5$.

\subsection{Training}

The predictor follows the multi-task training setting of ComboNet~\cite{jin2021deeplearningcovid}, jointly optimizing three objectives: drug--target interaction prediction, single-agent antiviral activity prediction, and drug-combination synergy prediction. These auxiliary objectives help address the limited number of SARS-CoV-2 combination training pairs (88 pairs after deduplication) by sharing molecular representations across related tasks.

\subsection{Mask-Aware Calibration}

After predictor training, all pretrained predictor parameters are frozen. A one-time mask-aware calibration then learns only the neutral mask embedding $e_{\mathrm{mask}} \in \mathbb{R}^{d_h}$ and the local reconditioning operator used during perturbation-based validation; the predictor weights remain unchanged throughout. During calibration, randomly sampled connected molecular regions are replaced by $e_{\mathrm{mask}}$ and the original prediction objectives are used to calibrate these perturbation-interface components. This step reduces the distribution shift that would otherwise occur when motif atoms are masked. After calibration, $e_{\mathrm{mask}}$ and the reconditioning operator are frozen together with the predictor for all \methodname{} explanations.

\subsection{Predictive Adequacy}

Table~\ref{tab:predictor_comparison} compares the fixed predictor against published baselines on the SARS-CoV-2 test set. The predictor achieves a test ROC-AUC of 0.85, providing a sufficiently informative fixed target for post-training explanation. Predictor accuracy is not a contribution of this work; the comparison is included solely to verify that the explanation target is meaningful.

\begin{table}[!htbp]
\centering
\caption{Predictor adequacy: test ROC-AUC on the SARS-CoV-2 combination benchmark.}
\label{tab:predictor_comparison}
\small
\begin{tabular}{lc}
\toprule
Model & Test ROC-AUC \\
\midrule
Random Forest & 0.62 \\
DeepSynergy & 0.68 \\
DeepDDS & 0.80 \\
ComboNet (original) & 0.82 \\
Our predictor (fixed target) & 0.85 \\
\bottomrule
\end{tabular}
\end{table}

\end{document}